\def\year{2020}\relax
\documentclass[letterpaper]{article} 
\usepackage{aaai20}  
\usepackage{times}  
\usepackage{helvet} 
\usepackage{courier}  
\usepackage[hyphens]{url}  
\usepackage{graphicx} 
\usepackage{graphicx}  
\usepackage{microtype}
\usepackage{booktabs}
\usepackage{makecell}
\usepackage{tabularx}
\usepackage{colortbl}
\usepackage{amsthm}
\usepackage{amssymb}
\usepackage{amsmath}
\usepackage{amsfonts}
\usepackage{bm}
\usepackage{arydshln}
\usepackage{multirow}
\definecolor{Gray}{gray}{0.95}

\newcommand{\citet}[1]{\citeauthor{#1} \shortcite{#1}}
\newcommand{\citep}{\cite}

\usepackage{dcolumn}
\newcolumntype{d}[1]{D{.}{.}{#1}}
\makeatletter
\newcolumntype{B}[3]{>{\boldmath\DC@{#1}{#2}{#3}}c<{\DC@end}}
\makeatother
\nocopyright

\title{A General Framework for Implicit and Explicit Debiasing \\ of Distributional Word Vector Spaces}

\author{\Large \textbf{Anne Lauscher,\textsuperscript{\rm 1} Goran Glava\v{s},\textsuperscript{\rm 1} Simone Paolo Ponzetto,\textsuperscript{\rm 1} and Ivan Vuli\'{c}\textsuperscript{\rm 2}}\\ 
\textsuperscript{\rm 1}Data and Web Science Research Group\\
University of Mannheim \\
\{anne, goran, simone\}@informatik.uni-mannheim.de \\
\textsuperscript{\rm 2} Language Technology Lab \\
University of Cambridge \\
iv250@cam.ac.uk
}

\begin{document}

\maketitle

\begin{abstract}
 Distributional word vectors have recently been shown to encode many of the human biases, most notably gender and racial biases, and models for attenuating such biases have consequently been proposed. However, existing models and studies (1) operate on under-specified and mutually differing bias definitions, (2) are tailored for a particular bias (e.g., gender bias) and (3) have been evaluated inconsistently and non-rigorously. In this work, we introduce a general framework for debiasing word embeddings. We operationalize the definition of a bias by discerning two types of bias specification: explicit and implicit. We then propose three debiasing models that operate on explicit or implicit bias specifications and that can be composed towards more robust debiasing. Finally, we devise a full-fledged evaluation framework in which we couple existing bias metrics with newly proposed ones. Experimental findings across three embedding methods suggest that the proposed debiasing models are robust and widely applicable: they often completely remove the bias both implicitly and explicitly without degradation of semantic information encoded in any of the input distributional spaces. Moreover, we successfully transfer debiasing models, by means of cross-lingual embedding spaces, and remove or attenuate biases in distributional word vector spaces of languages that lack readily available bias specifications.     
  
\end{abstract}

\definecolor{Gray}{gray}{0.95}

\section{Introduction}

Distributional word vectors have been recently shown to encode prominent human biases related to, e.g., gender or race \cite{Bolukbasi:2016:MCP:3157382.3157584,Caliskan183,manzini2019black}. Such biases are observed across languages and embedding methods \cite{lauscher2019we}, both in static and contextualized word embeddings \cite{zhao2019gender}. While this issue requires remedy, the finding itself is hardly surprising: we project our biases, in terms of biased word co-occurrences, into the texts we produce. Consequently, this is propagated to embedding models, both static \cite{mikolov2013distributed,pennington2014glove,Bojanowski:2017tacl}  and contextualized~\cite{peters2018deep} alike, by virtue of the distributional hypothesis~\cite{Harris:1954}.\footnote{Borrowing the famous example \cite{Bolukbasi:2016:MCP:3157382.3157584}, \textit{man} will be found more often in the same context with \textit{programmer}, and \textit{woman} with \textit{homemaker} in any sufficiently large corpus.} While biases may be useful for diachronic or sociological analyses \cite{GargE3635}, they (1) raise ethical issues, since biases are amplified by machine learning models using embeddings as input \cite{zhao2017men}, and (2) impede tasks like coreference resolution \cite{zhao2018gender,Rudinger2018gender} and abusive language detection \cite{park-etal-2018-reducing}. 
%
%

A number of methods for attenuating and eliminating human-like biases in word vectors have been proposed recently \cite{Bolukbasi:2016:MCP:3157382.3157584,zhao2018gender,zhao-etal-2018-learning,dev2019attenuating}. While they address the same types of bias -- primarily the gender bias -- they start from different bias ``specifications'' and either lack proper empirical evaluation \cite{Bolukbasi:2016:MCP:3157382.3157584} or employ different evaluation procedures, both hindering a direct comparison of the methods' ``debiasing abilities'' \cite{zhao2019gender,dev2019attenuating,manzini2019black}. What is more, the most prominent debiasing models \cite{Bolukbasi:2016:MCP:3157382.3157584,zhao-etal-2018-learning} have been criticized for merely masking the bias instead of removing it \cite{gonen2019lipstick}. 
To resolve inconsistencies in the current debiasing research and evaluation, in this work we propose a general debiasing framework DEBIE (\textbf{DEB}iasing embeddings \textbf{I}mplicitly and \textbf{E}xplicitly), which operationalizes bias specifications, groups models according to the bias specification type they operate on, and evaluates models' abilities to remove biases both explicitly and implicitly \cite{gonen2019lipstick}. 



We first define two types of bias specifications -- \textit{implicit} and \textit{explicit} -- and propose a method of augmenting bias specifications with the help of embeddings specialized for semantic similarity \cite{Mrksic:2017tacl,ponti2018adversarial}. We then introduce the main contributions of this work as follows. First, we present three novel debiasing models. (1) We adjust the linear projection method of \citet{dev2019attenuating}, an extension of the debiasing model of \citet{Bolukbasi:2016:MCP:3157382.3157584}, to operate on the augmented implicit bias specifications. (2) We then propose an alternative model that projects the embedding space to itself using the term sets from implicit bias specification as the projection signal. (3) Finally, we propose a simple and effective neural debiasing model, which is, to the best of our knowledge, the first debiasing model that operates on an explicit bias specification. All three models perform \textit{post-hoc} debiasing: they can be applied to any pretrained distributional word vector space.\footnote{In contrast, debiasing models like GN-GloVe \cite{zhao-etal-2018-learning} integrate debiasing constraints into objectives of embedding models like GloVe \cite{pennington2014glove}, and thus cannot be directly ported to other embedding models.} As another contribution, we combine existing bias metrics with newly proposed ones and assemble an evaluation suite that tests word vectors for explicit biases, implicit biases, and (preservation of) semantic quality. Finally, by coupling the proposed debiasing models with the cross-lingual embedding spaces \cite{ruder2017survey,glavas2019properly}, we facilitate cross-lingual debiasing transfer: we successfully debias embedding spaces in target languages without bias specifications in those languages. We hope that our work will lead to standardization of preprocessing and evaluation procedures in debiasing research and to increased comparability of debiasing models.\footnote{The code is available at \url{https://github.com/umanlp/DEBIE}.}


\section{General Debiasing Framework}
In what follows, we first formalize two bias specifications -- implicit and explicit. We then introduce new debiasing models: two operate on the implicit bias specification and the third on the explicit bias specification. Finally, we show how to debias word embeddings in a variety of target languages via cross-lingual embeddings.





\subsection{Bias Specifications} 
\label{sec:spec}

An \textit{implicit bias specification} $B_I = (T_1, T_2)$ consists of two sets of \textit{target} terms with respect to which a bias is expected to exist in the embedding space. For example, two sets of science and art terms, $T_1$ = $\{\textit{physics}, \textit{chemistry}, \textit{experiment}\}$ and $T_2 = \{\textit{poetry}, \textit{dance}, \textit{drama}\}$ constitute an implicit specification of the gender bias.
Strictly speaking, $B_I$ does not specify a bias directly -- it merely specifies two categories of concepts for which we \textit{implicitly} assume that there exists some set of reference terms $A$ (e.g., male terms \textit{man}, \textit{father} and/or female terms like \textit{woman}, \textit{girl}) with respect to which $T_1$ and $T_2$ exhibit differences. Most existing debiasing models \cite{Bolukbasi:2016:MCP:3157382.3157584,zhao-etal-2018-learning,dev2019attenuating,manzini2019black} operate on $B_I = (T_1, T_2)$, i.e., not requiring reference terms $A$. 


An \textit{explicit bias specification} $B_E$ defines, in addition to sets $T_1$ and $T_2$, one or more reference \textit{attribute} sets. We consider an explicit bias specification with a single attribute set, $B_E = (T_1, T_2, A)$ (as employed by our \textsc{DebiasNet} model),\footnote{The attribute set $A$ can be any set of attributes towards which the bias is to be removed. In our experiments, we joined the WEAT test specification attribute sets $A_1$ and $A_2$.} and also with two (opposing) attribute sets, $B_E = (T_1, T_2, A_1, A_2)$, as used in WEAT tests \cite{Caliskan183}.



\setlength{\tabcolsep}{12pt}
\def\arraystretch{0.89}
\begin{table*}[t!]
\centering
{\small
\begin{tabularx}{\linewidth}{l lX}
\toprule
Initial $T_1$ & \textit{science}, \textit{technology}, \textit{physics}, \textit{chemistry}, \textit{Einstein}, \textit{NASA}, \textit{experiment}, \textit{astronomy}\\ 
Initial $T_2$ & \textit{poetry}, \textit{art}, \textit{Shakespeare}, \textit{dance}, \textit{literature}, \textit{novel}, \textit{symphony},  \textit{drama}\\ 
Initial $A_1$ & \textit{brother}, \textit{father}, \textit{uncle}, \textit{grandfather}, \textit{son}, \textit{he}, \textit{his}, \textit{him}\\
Initial $A_2$ & \textit{sister}, \textit{mother}, \textit{aunt}, \textit{grandmother}, \textit{daughter}, \textit{she}, \textit{hers}, \textit{her}\\ 
\midrule
Augmentation $T_1$ & \makecell[l]{\textit{automation}, \textit{radiochemistry}, \textit{test}, \textit{biophysics}, \textit{learning}, \textit{electrodynamics}, \textit{biochemistry}, \textit{astrophysics}, \textit{astrometry}} \\ 
Augmentation $T_1$ & \makecell[l]{\textit{orchestra}, \textit{artistry}, \textit{dramaturgy}, \textit{poesy}, \textit{philharmonic}, \textit{craft}, \textit{untried}, \textit{hop}, \textit{poem}, \textit{dancing}, \textit{dissertation}, \textit{treatise}} \\
Augmentation $A_1$ & \textit{beget}, \textit{buddy}, \textit{forefather}, \textit{man}, \textit{nephew}, \textit{own}, \textit{himself}, \textit{theirs}, \textit{boy}, \textit{crony}, \textit{cousin}, \textit{grandpa}, \textit{granddad} \\ 
Augmentation $A_2$ & \textit{niece}, \textit{girl}, \textit{parent}, \textit{grandma}, \textit{granny}, \textit{woman}, \textit{theirs}, \textit{sire}, \textit{auntie}, \textit{sibling}, \textit{herself}, \textit{jealously}, \textit{stepmother}, \textit{wife} \\ 
\bottomrule
\end{tabularx}
}
\caption{Initial and augmented gender bias specifications. Test T8 from WEAT \cite{Caliskan183}.}
\label{tbl:augmentation}
\end{table*}

\paragraph{Augmentation of Bias Specifications.}
The initial bias specification ($B_I$ or $B_E$) commonly contains only a handful of words in each target and attribute set. These are commonly the most representative words of a category (e.g., \textit{man}, \textit{boy}, \textit{father} to represent the category \textit{male}). However, in order to provide a finer-grained bias specification, we propose to augment each term set with synonyms and semantically similar words of the initial terms. We therefore extract nearest neighbours of initial terms from an embedding space specialized to accentuate true semantic similarity and attenuate other types of semantic association \cite[\textit{inter alia}]{Faruqui:2015naacl,vulic-etal-2018-post,glavavs2018explicit}. For the augmentation process we rely on the recent state-of-the-art similarity specialization method of \citet{ponti2018adversarial}: for more details see the original work.


Given $B_I$ or $B_E$ and a similarity-specialized word vector space $\mathbf{X_{sim}}$, we augment each of the term sets in the specification by retrieving the top $k$ most (cosine-)similar terms from $\mathbf{X_{sim}}$ for each of the initial terms.\footnote{We discard nearest neighbors initially present in other sets of the same bias specification: e.g., if we retrieve an augmentation candidate \textit{woman} for an initial $T_1$ term \textit{man}, \textit{woman} will not be added to $T_1$ if it exists in $T_2$ (or in $A$-s).} Extending bias specification sets using a similarity-specialized word vector space -- as opposed to a regular distributional space -- reduces the noisy augmentation stemming from semantic relatedness instead of true semantic similarity.\footnote{We also considered using clean lexical knowledge from WordNet \citep{Miller:1995:WLD:219717.219748} directly, but this resulted in much lower recall as well as less accurate augmentation candidates.} 
%
Table~\ref{tbl:augmentation} illustrates the initial bias specification and the corresponding augmentation (showing $k=2$ nearest neighbors, without the initial terms) for one explicitly defined gender bias.

\subsection{Debiasing Models}
\label{sec:models}

We present three debiasing models, two of which operate on $B_I = (T_1, T_2)$ and one on the explicit bias specification $B_E = (T_1, T_2, A)$. 

\paragraph{Generalized Bias-Direction Debiasing (\textsc{GBDD})} focuses on $B_I$ as a generalization of the linear projection model proposed by \citet{dev2019attenuating}, itself, in turn, an extension of the hard-debiasing model of \citet{Bolukbasi:2016:MCP:3157382.3157584}. 

The model of \citet{dev2019attenuating} requires a stricter bias specification than our $B_I$: it requires $T_1$ and $T_2$ to be ordered lists of equal length $L$, so that the so-called equivalence pairs $\{(t^l_1, t^l_2)\}^L_{l = 1}$ can be created. For instance, $T_1 = $\{\textit{man}, \textit{father}, \textit{boy}\} and $T_2 = $\{\textit{woman}, \textit{mother}, \textit{girl}\} give rise to the following equivalence pairs: (\textit{man}, \textit{woman}), (\textit{father}, \textit{mother}), and (\textit{boy}, \textit{girl}). For each equivalence pair $(t^l_1, t^l_2)$ they compute the \textit{bias direction vector} $\mathbf{b_l}$ by subtracting the vector of term $t^l_2$ from the vector of term $t^l_1$. We find this bias specification overly restrictive: it requires an additional effort to create true equivalence pairs from $T_1$ and $T_2$ \textit{and} it produces only $L$ partial bias direction vectors. In contrast, we propose to create one bias direction vector $\mathbf{b_{ij}}$ for each pair $(t^i_1, t^j_2)$, $t^i_1 \in T_1$, $t^j_2 \in T_2$. If $T_1$ and $T_2$ truly specify categories that are opposite in some regard (e.g., gender-wise), then any pair $(t^i_1, t^j_2)$  should induce a meaningful partial bias direction vector. This way we also obtain a much larger number of partial bias direction vectors (e.g., $L^2$ if $T_1$ and $T_2$ are of the same length $L$): this should result in a more reliable \textit{general bias direction vector}, computed as follows.
%
%
We stack all of the obtained bias direction vectors $\mathbf{b_{ij}}$ corresponding to pairs $(t^i_1, t^j_2)$, $t^i_1 \in T_1$, $t^j_2 \in T_2$ to form a bias direction matrix $\bm{B}$. We then obtain the \textit{global} bias direction vector $\mathbf{b}$ as the top singular vector of $\bm{B}$, i.e., as the first row of matrix $\bm{V}$, where $\bm{U\Sigma V^\top}$ is the singular value decomposition of $\bm{B}$. Let $\mathbf{x}$ be the $\ell_2$-normalized $d$-dimensional vector from a biased input vector space. Its debiased version is then computed as: 
{
\begin{equation}
    \text{GBDD}(\mathbf{x}) = \mathbf{x} - \langle \mathbf{x}, \mathbf{b} \rangle \mathbf{b}  
\end{equation}}%
\noindent where $\langle \cdot, \cdot \rangle$ denotes a dot product. In other words, the closer the vector $\mathbf{x}$ is to the global bias direction $\mathbf{b}$, the more it is bias-corrected (i.e., the larger portion of $\mathbf{b}$ is subtracted from $\mathbf{x}$). Vectors orthogonal to the bias direction $\mathbf{b}$ remain unchanged (zero dot-product with the bias vector $\mathbf{b}$).        

\paragraph{Bias-Alignment Model (BAM).} 
An alternative to computing a bias direction vector $\mathbf{b}$ is to use target-term pairs $(t^i_1, t^j_2)$, $t^i_1 \in T_1$, $t^j_2 \in T_2$ to learn a projection of the biased embedding space $\bm{X} \in \mathbb{R}^d$ to itself that (approximately) aligns $T_1$ and $T_2$. The idea behind this model stems from the research on projection-based cross-lingual word embeddings (CLWEs), where an orthogonal mapping between monolingual embedding spaces is learned from a set of word translations \cite{smith2017offline,glavas2019properly}.\footnote{Note that a self-consistent linear mapping $W$ is the one offering consistent mapping from one space to the other and back, $x = \bm{W}^\top\bm{W}x$ , i.e., $\bm{W}^\top\bm{W} = \bm{I}$, thus $W$ is orthogonal; an orthogonal projection $W (\bm{X}' = \bm{W}\bm{X})$ preserves all distances in $\bm{X}$, making $\bm{X}'$ isomorphic to $\bm{X}$.} 


Here, we use pairs $(t^i_1, t^j_2)$ to learn the debiasing projection of $\bm{X}$ with respect to itself. Let $\bm{X}_{T_1}$ and $\bm{X}_{T_2}$ be the matrices obtained by stacking (biased) vectors of left and right words of pairs $(t^i_1, t^j_2)$, respectively. We then learn the orthogonal map $\bm{W_X} = \bm{U}\bm{V}^\top$, where $\bm{U}\bm{\Sigma}\bm{V}^\top$ is the singular value decomposition of $\bm{X}_{T_2}\bm{X}_{T_1}^\top$. Since $\bm{W_X}$ is orthogonal, the projection $\bm{X}' = \bm{X}\bm{W_X}$ is isomorphic to the original space $\bm{X}$, and thus equally biased. However, the transformation (specified by $\bm{W_X}$) defines the angle and direction of debiasing. We obtain the debiased space by averaging the original space $\bm{X}$ and the projected space $\bm{X}'$:
%
{
\begin{equation}
    \text{BAM}(\mathbf{X}) = \frac{1}{2}\left(\mathbf{X} + \mathbf{X}\bm{W}_X\right).
\end{equation}}%

\paragraph{Explicit Neural Debiasing (\textsc{DebiasNet}).} 
The final model, dubbed \textsc{DebiasNet}, is a neural model that operates on the explicit bias specification $B_E$. It is inspired by the work on semantic specialization of word embeddings \cite{vulic-etal-2018-post,glavavs2018explicit}: but instead of using linguistic constraints (e.g., synonyms), we ``specialize'' the vector space by leveraging debiasing constraints. 
%

Given a biased input space $\bm{X}$ and the specification $B_E=(T_1, T_2, A)$, we learn a debiasing function $\text{DBN}(\bm{X};\mathbf{\theta})$ that transforms $\bm{X}$ to a debiased space $\bm{X}'$. We aim for the terms from both sets $T_1$ and $T_2$ to be similarly close to the terms from $A$ in $\bm{X}'$. 
%
For simplicity, we execute DBN$(\bm{X};\mathbf{\theta})$ as a feed-forward neural network with non-linear activations. The training set for learning the parameters $\theta$ consists of triples $(t_1 \in T_1, t_2 \in T_2, a \in A)$. It is obtained as a full Cartesian product $T_1 \times T_2 \times A$. Let $\mathbf{t}_1$, $\mathbf{t}_2$ and $\mathbf{a}$ be the respective vectors of $t_1$, $t_2$, and $a$ from the input biased space $\bm{X}$, and let $\mathbf{t}'_1$, $\mathbf{t}'_2$ and $\mathbf{a}'$ be their ``debiased'' transformations: $\mathbf{t}'_1 = \text{DBN}(\mathbf{t}_1; \theta)$, $\mathbf{t}'_2 = \text{DBN}(\mathbf{t}_2; \theta)$, and $\mathbf{a}' = \text{DBN}(\mathbf{a}; \theta)$.    
%
For a training instance $(t_1, t_2, a)$, we then minimize the following loss function $L_D$:  %
%
%
{
\begin{equation}
    L_D = \left(\cos_d\left(\mathbf{t}'_1, \mathbf{a}'\right) - \cos_d\left(\mathbf{t}'_2, \mathbf{a}'\right) \right)^2.
\end{equation}}%

\noindent $\cos_d(\cdot,\cdot)$ refers to the cosine distance. The objective pushes the terms from the two target sets $T_1$ and $T_2$ to be equidistant to the terms from the attribute set $A$. That is, it is designed to specifically remove the explicit bias. By minimizing $L_D$ as the only objective, the model would remove the bias, but it would also destroy the useful semantic information in the input space. We thus couple the objective $L_D$ with the regularization $L_R$ that prevents the debiased vectors to deviate too much from their original estimates:
%
{
\begin{equation}
    L_R\hspace{-0.2em}=\hspace{-0.2em}\cos_d\hspace{-0.2em}\left(\mathbf{t}_1, \mathbf{t}'_1\right) \hspace{-0.05em} + \hspace{-0.05em} \cos_d\hspace{-0.2em}\left(\mathbf{t}_2, \mathbf{t}'_2\right) \hspace{-0.05em} + \hspace{-0.05em} 
    \cos_d\hspace{-0.2em}\left(\mathbf{a}, \mathbf{a}'\right)
\end{equation}}%
\noindent The final loss is then $J = L_D + \lambda L_R$, with $\lambda$ as the regularization weight. The learned function is then applied to the full input space: $\bm{X}' = \text{DBN}(\bm{X}; \theta)$. 

%


\paragraph{Composing Debiasing Models.}
The presented models can be seamlessly composed with one another. For example, given an explicit specification $B_E$, we can first explicitly debias a distributional space $\mathbf{X}$ using \textsc{DebiasNet}. We can then apply either GBDD or BAM on the resulting vector space by deriving $B_I$ from $B_E$ (i.e., by considering only $T_1$ and $T_2$): e.g., $\bm{X}' = \text{GBDD}(\text{DBN}(\bm{X}))$. 


\subsection{Cross-Lingual Transfer of Debiasing} 
\label{sec:xling}
Cross-lingual word embeddings have been shown to be a viable solution for zero-shot language transfer of NLP models \cite{ruder2017survey,glavas2019properly}. Conceptually, given a source language $L_1$ with its monolingual distributional space $\bm{X}_{L1}$ and a target language $L_2$ with the space $\bm{X}_{L2}$, we can apply any $L1$ model trained on $\bm{X}_{L1}$ on the instances from $L2$, given a matrix $\bm{W}_{CL}$ that projects $\bm{X}_{L2}$ to $\bm{X}_{L1}$. From the plethora of cross-lingual word embedding models \cite[\textit{inter alia}]{smith2017offline,conneau2018word,artetxe2018robust}, we opt for a supervised projection-based model \cite{smith2017offline} that obtains $\bm{W}_{CL}$ by solving the Procrustes problem \cite{schonemann1966procrustes} on the set of word translation pairs.\footnote{Note that we obtain the cross-lingual projection $\bm{W}_{CL}$ in the similar way as debiasing projection $\bm{W}_X$ in BAM
; but now the aligned matrices contain vectors (each from respective language) corresponding to word translation pairs (not pairs created from bias target sets as in BAM).} We select this approach due to its simplicity and competitive zero-shot language transfer performance on other NLP tasks \cite{glavas2019properly}. With the cross-lingual projection matrix $\bm{W}_{CL}$ in place, the debiasing of the space $\bm{X}_{L2}$ amounts to composing the projection with the debiasing model in $L1$: e.g., for GBDD, $\bm{X}'_{L2} = \text{GBDD}_{L1}(\bm{X}_{L2}\bm{W}_{CL})$.


\section{Evaluation and Experimental Setup}

We now introduce the metrics for testing different aspects of debiased embedding spaces, and then outline two datasets used in our experiments. 

\subsection{Evaluation Aspects}
\label{sec:eb_eval}

We use three diverse tests to measure the presence of explicit bias, and two tests that focus on the presence of implicit bias. Finally, we test the debiased spaces for their ability to preserve the initial semantic information. 

\paragraph{Word Embedding Association Test (WEAT).} 

Introduced by \citet{Caliskan183}, WEAT tests the embedding space for the presence of an explicit bias defined as $B_E$=$(T_1, T_2, A_1, A_2)$. It computes the differential association between $T_1$ and $T_2$ based on their mean similarity with terms from the attribute sets $A_1$ and $A_2$: 
%
{
\begin{equation}
        s(B_E) = \sum_{t_1 \in T_1}{s(t_1, A_1, A_2)} - \sum_{t_2 \in T_2}{s(t_2, A1, A2)}\,
\end{equation}}%

\noindent The association $s$ of term $t\in T_i$ is computed as: 
%
{
\begin{equation}
    s(t,\hspace{-0.2em}A_1\hspace{-0.2em},\hspace{-0.2em}A_2)\hspace{-0.2em}=\hspace{-0.2em} \frac{1}{|A_1|}\hspace{-0.5em}\sum_{a_1 \in A_1}{\hspace{-0.7em}\cos(\mathbf{t}, \mathbf{a_1})}  -  \frac{1}{|A_2|}\hspace{-0.5em}\sum_{a_2 \in A_2}{\hspace{-0.7em}\cos(\mathbf{t}, \mathbf{a_2})} 
\end{equation}}%

\noindent The significance of the statistic is computed by comparing $s(B_E)$ with the scores $s(B^*_E)$ obtained with all permutations $B^*_E = (T^*_1, T^*_2, A_1, A_2)$, where $T^*_1$ and $T^*_2$ are equally sized partitions of $T_1 \cup T_2$. The $p$-value of the test is the probability of $s(B^*_E) > s(B_E)$. The ``amount'' of bias, the so-called \textit{effect size}, is then a normalized measure of separation between association distributions:
%
{
\begin{equation}
\frac{\mu\hspace{-0.1em}\left(\{s(t_1, A_1, A_2)\}_{t_1 \in T_1}\right) - \mu\hspace{-0.1em}\left(\{s(t_2, A_1, A_2)\}_{t_2 \in T_2}\right)}{\sigma\left(\{s(t, A_1, A_2)\}_{t \in T_1 \cup T_2}\right)}
\end{equation}}%

\noindent where $\mu$ is the mean and $\sigma$ is the standard deviation.

\paragraph{Embedding Coherence Test (ECT).} It quantifies the amount of explicit bias $B_E$=$\{T_1, T_2, A\}$ \cite{dev2019attenuating}. Unlike WEAT, 
it compares vectors of target sets $T_1$ and $T_2$ (averaged over the constituent terms) with vectors from a single attribute set $A$. ECT first computes the mean vectors for the target sets $T_1$ and $T_2$: $\mathbf{\mu}_1 = \frac{1}{|T_1|}\sum_{t_1 \in T_1}{\mathbf{t}_1}$ and $\hspace{0.5em} \mathbf{\mu}_2 = \frac{1}{|T_2|}\sum_{t_2 \in T_2}{\mathbf{t}_2}$. Next, for both $\mathbf{\mu}_1$ and $\mathbf{\mu}_1$ it computes the (cosine) similarities with vectors of all $\mathbf{a} \in A$. The two resultant vectors of similarity scores, $\mathbf{s}_1$ (for $T_1$) and $\mathbf{s}_2$ (for $T_2$) are used to obtain the final ECT score. It is the Spearman's rank correlation between the rank orders of $\mathbf{s}_1$ and $\mathbf{s}_2$ -- the higher the correlation, the lower the bias.    


\paragraph{Bias Analogy Test (BAT).} Based on the observation of \cite{Bolukbasi:2016:MCP:3157382.3157584} that in a biased vector space $\mathit{programmer} - \mathit{homemaker} \approx \mathit{man} - \mathit{woman}$, \citet{dev2019attenuating} proposed an analogy-based bias test: Embedding Quality Test (EQT). However, EQT depends on WordNet to extend the bias definition with synonyms and plurals of bias specification terms. In contrast, we propose an alternative Bias Analogy Test (BAT) that relies only on the specification $B_E = (T_1, T_2, A_1, A_2)$.

We first create all possible biased analogies $\mathbf{t}_1 - \mathbf{t}_2 \approx \mathbf{a}_1 - \mathbf{a}_2$ for $(t_1, t_2, a_1, a_2) \in T_1 \times T_2 \times A_1 \times A_2$. We then create two query vectors from each analogy: $\mathbf{q}_1 = \mathbf{t}_1 - \mathbf{t}_2 + \mathbf{a}_2$ and $\mathbf{q}_2 = \mathbf{a}_1 - \mathbf{t}_1 + \mathbf{t}_2$ for each 4-tuple $(t_1, t_2, a_1, a_2)$. We then rank the vectors in the vector space $\bm{X}$ according to the Euclidean distance with each of the query vectors. In a biased space, we expect the vector $\mathbf{a}_1$ to be ranked higher for the query $\mathbf{q}_1$ than the vectors of terms from the opposing attribute set $A_2$ (e.g., for a gender-biased space we expect \textit{woman} to be ranked higher than \textit{father} or \textit{boy} for the query \textit{man} - \textit{programmer} + \textit{homemaker}). Also, $\mathbf{a}_2$ is expected to be more similar to $\mathbf{q}_2$ than vectors of $A_1$ terms . The BAT score is the percentage of cases where: (1) $a_1$ is ranked higher than a term $a'_2 \in A_2\setminus\{a_2\}$ for $\mathbf{q}_1$ and (2) $a_2$ is ranked higher than a term $a'_1 \in A_1\setminus\{a_1\}$ for $\mathbf{q}_2$. 

\paragraph{Implicit Bias Tests.} \citet{gonen2019lipstick} recently suggested that the two sets of target terms can still be clearly distinguished (with KMeans clustering, or in a supervised manner with an SVM classifier) from one another after applying debiasing procedures of \cite{Bolukbasi:2016:MCP:3157382.3157584} and \cite{zhao-etal-2018-learning}. We adopt their approach and test the debiased spaces for the presence of implicit bias by clustering terms from $T_1$ and $T_2$ with KMeans++, and by classifying them using an SVM with the RBF kernel: it is trained on the vectors of terms from the augmentations of target sets. For each debiasing model, we average the clustering and classification scores over $20$ independent runs.

\paragraph{Semantic Quality.} Debiasing procedures change the topology of the input vector space; we thus have to verify that debiasing does not occur at the expense of the encoded semantic information. We test the debiased embedding spaces on two standard word similarity/relatedness benchmarks: SimLex-999 \cite{hilldoi:10.1162} and WordSim-353 \cite{finkelstein2002placing}.  

\subsection{Evaluation Datasets}

Our proposed framework is versatile as it enables debiasing models to operate on any bias specified in the $B_I$ or $B_E$ format. 
To demonstrate this, we evaluate the debiasing models from the previous section on two different bias specifications: tests T1 and T8 from the WEAT dataset \citep{Caliskan183}. WEAT tests are given as explicit bias specifications $B_E$= $(T_1, T_2, A_1, A_2)$. 

\paragraph{WEAT T8: Gender Bias Test.}
WEAT T8, shown in Table~\ref{tbl:augmentation}, encodes a type of a gender bias in relation to affinities towards science and art. $T_1$ contains terms from the areas of science and technology, whereas $T_2$ contains art terms. Attribute sets contain male ($A_1$) and female ($A_2$) terms. In a gender-biased vector space the scientific targets are expected to be more strongly associated with male attributes, and artistic targets with female terms.       


\paragraph{WEAT 1: Flowers vs. Insects.}
WEAT T1 specifies another bias type: the difference in \textit{sentiment} humans attach to \textit{insects} as opposed to \textit{flowers}. Target sets contain different flowers ($T_1$) and insect species ($T_2$), and attribute sets contain universally positive ($A_1$) and negative ($A_2$) terms. The full bias specification of WEAT T1 is available in the supplementary. 

\paragraph{XWEAT.} 
For evaluating the language transfer setup, we use bias specifications in target languages as our test data. We use tests T1 and T8 from XWEAT, created by \citet{lauscher2019we} by translating the English (\textsc{en}) WEAT tests to six languages: German (\textsc{de}), Spanish (\textsc{es}), Italian (\textsc{it}), Russian (\textsc{ru}), Croatian (\textsc{hr}), and Turkish (\textsc{tr}).   

\subsection{Preprocessing and Training Setup}
%

\noindent \textbf{Augmented Bias Specifications.} We first augment the bias specifications using a similarity-specialized embedding space produced by \citet{ponti2018adversarial}\footnote{Available at: \url{https://tinyurl.com/y273cuvk}.} based on the \textsc{en} fastText embeddings \cite{Bojanowski:2017tacl}. For WEAT T8, we augment the target and attribute lists with $k=4$ nearest neighbours of each term. As the initial lists of WEAT T1 are longer than those of T8, we use $k=2$ with T1. We train all debiasing models using bias specifications containing \textit{only} the augmentation terms (i.e., without the initial bias specification terms); we use the initial terms for testing.             

\paragraph{Input Word Embeddings.} We test the robustness of debiasing models on three different word embedding models trained on Wikipedia: CBOW \cite{mikolov2013distributed}, GloVe \cite{pennington2014glove}, and fastText (FT) \cite{Bojanowski:2017tacl}. For cross-lingual transfer, we induce a multilingual space spanning seven languages (\textsc{en} + 6 targets) by projecting FT vectors of each target to the \textsc{EN} space. Following an established procedure \cite{glavas2019properly}, we learn projections $\bm{W}_{CL}$ using automatically compiled translations of the 5K most frequent \textsc{en} words.


\paragraph{Training Setup.} For GBDD and BAM there is a deterministic closed-form solution for any given bias specification. On the other hand, the hyper-parameters of \textsc{DebiasNet} are optimized via grid search and cross-validation on the training set. The final \textsc{DebiasNet} model uses $5$ hidden layers with $300$ units each and the weight $\lambda$ is fixed to 0.2. 



\section{Results and Analysis}
We first report debiasing results on three \textsc{en} distributional spaces, for the individual models as well as for three composite models: GBDD $\circ$ BAM = GBDD(BAM($\bm{X}$)), BAM $\circ$ GBDD, and GBDD $\circ$ \textsc{DebiasNet}.\footnote{BAM and \textsc{DebiasNet} display similar results and so does their composition. For brevity, we thus omit the scores of BAM $\circ$ \textsc{DebiasNet}. We also do not report the scores with \textsc{DebiasNet} $\circ$ GDBB as its scores were similar to its inverse composition GDBB $\circ$ \textsc{DebiasNet} in our preliminary tests.} We then show the results for the cross-lingual debiasing transfer. Finally, we analyze the topology of debiased spaces.


\subsection{Main Evaluation}

\setlength{\tabcolsep}{4pt}
\begin{table*}[!t]
\def\arraystretch{0.85}
\centering
{
\begin{tabularx}{\linewidth}{l l ccc cc cc  ccc cc cc cc}
\toprule
& & \multicolumn{7}{c}{WEAT T8 (gender bias, science vs.~art)} & \multicolumn{7}{c}{WEAT T1 (sentiment, flowers vs. insects)} \\
\cmidrule(lr){3-9} \cmidrule(lr){10-16}
 & & \multicolumn{3}{c}{Explicit} & \multicolumn{2}{c}{Implicit} & \multicolumn{2}{c}{SemQ} & \multicolumn{3}{c}{Explicit} & \multicolumn{2}{c}{Implicit} & \multicolumn{2}{c}{SemQ} \\ 
\cmidrule(lr){3-5} \cmidrule(lr){6-7} \cmidrule(lr){8-9} \cmidrule(lr){10-12} \cmidrule(lr){13-14} \cmidrule(lr){15-16} 
\multicolumn{2}{c}{Model} & WEAT & ECT & BAT & KM & SVM & SL & WS & WEAT & ECT & BAT & KM & SVM & SL & WS \\ \midrule
\rowcolor{Gray}
\textbf{FT} & Distributional & 1.30 & 73.5 & 41.0 & 100 & 100 & 38.2 & 73.8 
& 1.67 & 46.2 & 56.1 & 95.7 & 100 & 38.2 & 73.0  \\ \hdashline
& GBDD & 0.96 & 84.7 & 33.9 & 62.9 & \textbf{50.0} & \textbf{38.4} & \textbf{73.8}
& 0.08* & 96.2 & 41.7 & 56.0 & 53.1 & 38.1 & 72.9  \\
\rowcolor{Gray}
& BAM & 0.10* & 71.8 & 38.4 & 99.8 & 100 & 37.7 & 70.4
& 1.57 & 50.3 & 56.0 & 95.7 & 100 & 37.4 & 71.5  \\
& DBN & \textbf{0.05}* & 79.1 & \textbf{33.6} & 99.8 & 100 & 34.1 & 65.1 
& 0.18* & 79.8 & 45 & 95.7 & 100 & 35.09 & 68.6 \\ \hdashline
\rowcolor{Gray}
& GBDD $\circ$ BAM & 0.18* & \textbf{94.4} & 38.7 & 65.1 & 65.3 & 37.7 & 70.2
& 0.42* & 89.3 & 48.1 & 75.0 & 91.4 & 37.3 & 71.3 \\
& BAM $\circ$ GBDD & 0.57* & 90.3 & 34.6 & \textbf{60.1} & \textbf{50.0} & 36.4 & 72.6 
& \textbf{0.07}* & 94.4 & 42.4 & 56.9 & \textbf{51.3} & 37.9 & 68.4 \\
\rowcolor{Gray}
& GBDD $\circ$ DBN & 0.11* & 81.5 & 37.4 & 65.8 & 50.3 & 33.9 & 64.6 
& -0.08* & 95.9 & \textbf{41.9} & \textbf{54.6} & \textbf{52.0} & 34.9 & 68.4 \\ \midrule
\textbf{CBOW} & Distributional & 0.81* & -24.0 & 45.6 & 90.6 & 93.4 & 34.7 & 59.4 
& 1.13 & 78.1 & 50.2 & 62.6 & 93.9 & \textbf{34.7} & \textbf{59.4} \\ \hdashline
\rowcolor{Gray}
& GBDD & 0.38* & 50.9 & 43.4 & 59.5 & \textbf{50.0} & \textbf{34.8} & \textbf{59.8} 
& -0.07* & 90.7 & \textbf{41.1} & 55.7 & 51.9 & \textbf{34.7} & \textbf{59.4} \\
& BAM &  0.14* & 36.8 & 51.1 & 95.1 & 89.4 & 33.4 & 59.2
& 0.44* & 82.4 & 50.7 & 60.9 & 94.4 & 34.4 & 59.3\\
\rowcolor{Gray}
& DBN & 0.45* & 4.7 & 57.5 & 97.4 & 98.4 & 33.9 & 52.2  
& 0.60 & 82.5 & 46 & 85.7 & 90.8 & 33.4 & 53.4 \\ \hdashline
& GBDD $\circ$ BAM & \textbf{0.00}* & \textbf{69.4} & 50.3 & \textbf{52.7} & 68.8 & 33.4 & 59.3  
& \textbf{-0.04}* & \textbf{91.3} & 48.7 & 60.7 & 68.1 & 34.5 & 59.2 \\
\rowcolor{Gray}
& BAM $\circ$ GBDD & 0.09* & 65.6 & 42.7 & 62.6 & \textbf{50.0} & 33.2 & 56.9  
& -0.17* & 89.2 & 45.3 & 55.6 & \textbf{51.1} & 33.2 & 57 \\
& GBDD $\circ$ DBN & 0.38* & -3.5 & 57.6 & 61.9 & 50.3 & 34.0 & 52.1  
& -0.15* & 90.5 & 41.3 & \textbf{55.4} & 52.6 & 33.4 & 53.3 \\ \midrule
%
\rowcolor{Gray}
\textbf{GloVe} & Distributional & 1.28 & 84.1 & 36.3 & 100 & 100 & \textbf{36.9} & \textbf{60.5} 
& 1.38 & 76.2 & 40.5 & 94.1 & 100 & \textbf{36.9} & 60.5 \\ \hdashline
& GBDD & 0.95 & 89.7 & 29.1 & 57.4 & 50.6 & \textbf{36.9} & 59.6
& 0.44* & \textbf{92.4} & 32.7 & 55.6 & 54.5 & 36.8 & \textbf{60.7} \\
\rowcolor{Gray}
& BAM & 1.08 & 89.7 & 27.8 & 96 & 100 & 36.2 & 59.5
& 0.96 & 82.1 & 39.2 & 90.7 & 100 & 34.4 & 56.4 \\
& DBN & 0.83* & 81.5 & 30.8 & 100 & 100 & 35.9 & 58.6 
& 0.55 & 77.6 & 34.8 & 95.3 & 100 & 36.7 & 59.1 \\ \hdashline
\rowcolor{Gray}
& GBDD $\circ$ BAM & 0.98 & 94.7 & \textbf{25.8} & 63.6 & 79.1 & 36.6 & 59.3 
& 0.40* & 90.7 & 36.5 & 57.7 & 76.5 & 34.2 & 56.4 \\
& BAM $\circ$ GBDD & 0.78* & 97.1 & 36.9 & \textbf{53.9} & \textbf{50.0} & 36.3 & 59.2
& 0.65 & 87.3 & 44.1 & \textbf{55.5} & \textbf{51.2} & 35.5 & 58.6 \\
\rowcolor{Gray}
& GBDD $\circ$ DBN & \textbf{0.51}* & \textbf{97.4} & 28.2 & 59.5 & \textbf{50.0} & 35.8 & 58.4
& \textbf{-0.03}* & 89.7 & \textbf{30.3} & 57.4 & 52.1 & 36.5 & 59.1 \\
\bottomrule
\end{tabularx}
}
\caption{Main results on two bias test sets, WEAT T8 and T1 for three \textsc{en} distributional spaces debiased with three models -- GBDD, BAM, and DebiasNet (DBN) -- and their compositions. We quantify the explicit bias (Explicit): WEAT, ECT, and BAT evaluation measures; implicit bias (Implicit): clustering with KMeans++ (KM) and classification with SVM; and the preservation of semantic quality (SemQ): word similarity scores on SimLex (SL) and WordSim-353 (WS). Asterisks (*) indicate insignificant ($\alpha = 0.05$) bias effects for the WEAT evaluation measure.}
\label{tbl:base_results}
\end{table*}
\setlength{\tabcolsep}{2.2pt}
\begin{table*}[t!]
\def\arraystretch{0.85}
\centering
{
\begin{tabularx}{\linewidth}{l l ccc ccc ccc ccc ccc ccc}
\toprule
& & \multicolumn{3}{c}{\textsc{de}} & \multicolumn{3}{c}{\textsc{es}} & \multicolumn{3}{c}{\textsc{it}} & \multicolumn{3}{c}{\textsc{ru}} & \multicolumn{3}{c}{\textsc{hr}} & \multicolumn{3}{c}{\textsc{tr}} \\ 
\cmidrule(lr){3-5} \cmidrule(lr){6-8} \cmidrule(lr){9-11} \cmidrule(lr){12-14} \cmidrule(lr){15-17} \cmidrule(lr){18-20} 
\multicolumn{2}{c}{Model} & W & KM & SL & W & KM & SL & W & KM & SL & W & KM & SL & W & KM & SL & W & KM & SL \\ \midrule
\rowcolor{Gray}
\textbf{FT} & Distributional & \textbf{0.05}* & 98.3 & 40.7 & 1.16 & 99.8 & -- & 0.10* & 99.8 & \textbf{29.8} & 0.37* & 62 & 25.6 & 0.13* & 98.6 & \textbf{32.7} & 1.72 & 99.3 & -- \\ \hdashline
& GBDD & 0.15* & 55.4 & 40.7 & 0.41* & 60 & -- & -0.28* & \textbf{56.1} & \textbf{29.8} & 0.73* & 62.4 & \textbf{25.8} & 0.54* & \textbf{59.9} & 32.5 & 1.41 & 64.3 & -- \\
\rowcolor{Gray}
& BAM & -0.97 & 97.4 & 40.7 & 0.11* & 99.0 & -- & -0.70* & 99.6 & 29 & -0.41* & 74.4 & 25.1 & \textbf{-0.01}* & 93.5 & 32 & 1.49 & 98.8 & -- \\
& DBN & -0.15* & 97.4 & 36.2 & 0.76* & 100 & -- & -1.05 & 100 & 25.4 & \textbf{0.31}* & 77.9 & 20.7 & 0.25* & 99.9 & 25.3 & 1.54 & 100 & -- \\ \hdashline
\rowcolor{Gray}
& GBDD $\circ$ BAM & 0.35* & 57.6 & 35.9 & 0.78* & \textbf{52.4} & -- & -0.64* & 60.1 & 25.0 & 0.77* & 61.9 & 20.7 & 0.67* & 67.5 & 25.1 & 1.29 & 62.5 & -- \\
& BAM $\circ$ GBDD & -0.12* & 56.3 & \textbf{40.8} & \textbf{0.05}* & 58 & -- & -0.62* & 57.9 & 29 & 0.34* & \textbf{56.8} & 24.8 & 0.52* & 60.8 & 31.7 & \textbf{0.99} & \textbf{56.9} & -- \\
\rowcolor{Gray}
& GBDD $\circ$ DBN & -0.09* & \textbf{54.4} & 37.3 & 0.11* & 56.6 & -- & \textbf{-0.05}* & 58.9 & 27.1 & 0.59* & 61.6 & 25.4 & 0.68* & 75.4 & 29.4 & 1.27 & 62.4 & -- \\
\bottomrule
\end{tabularx}
}
\caption{Results for cross-lingual debiasing transfer on XWEAT T8 for six languages: \textsc{de}, \textsc{es}, \textsc{it}, \textsc{ru}, \textsc{hr}, and \textsc{tr}. Input word embeddings are fastText (FT) for all target languages.W=WEAT; KM=KMeans++; SL=SimLex.}
\label{tbl:xling}
\end{table*}


\begin{figure*}[!t]
    \centering
     \includegraphics[scale=0.38]{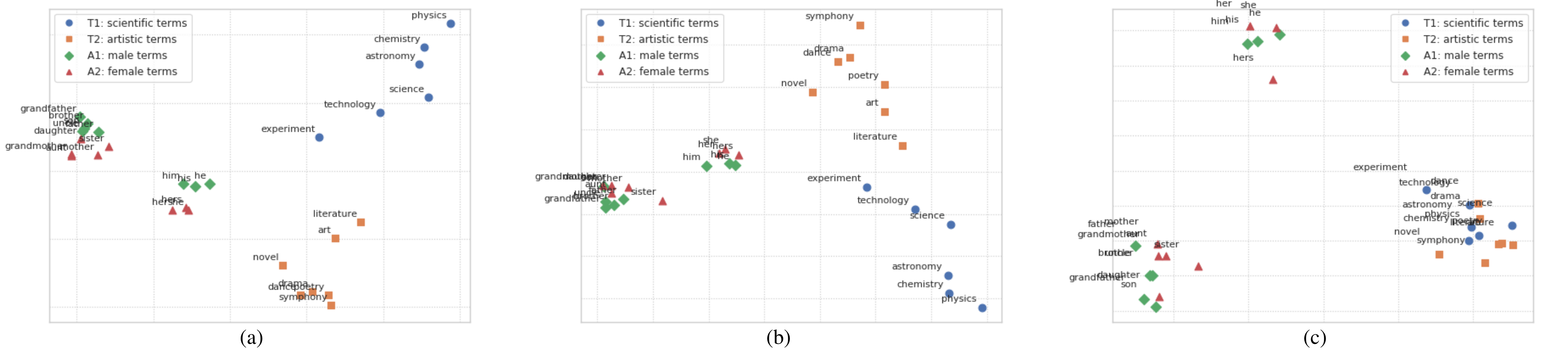}
    \caption{The topology of a vector space before and after debiasing.  Terms from WEAT T8 test: $T_1$ -- \textit{science} terms (blue), $T_2$ -- \textit{art} terms (orange), $A_1$ -- \textit{male} terms (green), and $A_2$ -- \textit{femal}e terms (red). 2D projection with PCA. (a) Distributional \textsc{en} FT vectors; (b) Debiased using BAM; (c) Debiased using GBDD.}
    \label{fig:topology}
\end{figure*}

\noindent \textbf{Biases of Distributional Spaces.} 
The main results are summarized in Table~\ref{tbl:base_results}. All three input distributional spaces generally exhibit explicit and implicit biases, with CBOW spaces displaying the lowest biases, both according to the WEAT tests (e.g., the effect size is even insignificant with $p < 0.05$ for the gender bias test T8) and the implicit bias tests of \citet{gonen2019lipstick}. Interestingly -- according to our BAT test, and despite the original claims and examples from \citet{Bolukbasi:2016:MCP:3157382.3157584} -- the encoded biases do not reflect strongly in the analogy tests. Nonetheless, our debiasing methods in most test settings manage to affect the input vector spaces by further reducing BAT scores. 



\paragraph{Comparison of Debiasing Models.} 
While the results vary across the two WEAT tests and evaluation metrics, GBDD emerges as the most robust model on average. It attenuates the explicit bias while being the most successful in removing the bias implicitly: the spaces debiased with GBDD completely confuse the KM clustering and SVM classifier. It also fully retains the useful semantic information: we do not observe drops on SL and WS compared to the input distributional spaces. 
While GBDD outperforms BAM and \textsc{DebiasNet} (DBN) on average according to ECT and BAT measures, it is not able to fully remove the explicit gender bias (T8) according to the WEAT test. 

Despite operating on an implicit specification $B_I$, BAM removes the explicit biases much better than the implicit ones. DBN seems even better than BAM in removing the explicit biases. This is not a surprise, since DBN is trained on an explicit bias specification. However both DBN and BAM are unsuccessful in removing the implicit biases. Moreover, DBN distorts the input space more than BAM, yielding substantial drops on SL and WS.

The complementarity of debiasing effects between GBDD, and BAM/DBN are confirmed by the performance of their compositions. All composition models robustly remove both explicit and implicit biases, also showing that there is no ``one model rules them all'' solution to various debiasing aspects. GBDD $\circ$ DBN most effectively removes the biases, but it inherits the undesirable semantic distortions of DBN. On the other hand, BAM $\circ$ GBDD offers solid bias removal while for the most part retaining the semantic quality of the space. 

\paragraph{Differences between Evaluation Measures.} 
The three aspects of evaluation complement each other: they all inform the selection of the most appropriate debiasing model w.r.t. the desired application-specific criteria.\footnote{E.g., Note that for some bias specifications, one might not want to reduce/remove the implicit bias. WEAT T1 can be seen as an example of such bias: while we may want to make \textit{insects} similarly \textit{good}/\textit{bad} as \textit{flowers}, we do not want to make them indistinguishable from \textit{flowers} in the vector space.} However, results of WEAT, ECT, and BAT are not always aligned. For example, the CBOW space is unbiased according to the WEAT test, but extremely biased (negative correlation!) according to ECT. In contrast, GloVe vectors are biased according to WEAT but not according to ECT (correlation of $0.84$). These findings point to different bias aspects, accentuating the need for multiple, mutually complementary, bias measures.   

%
%

\subsection{Cross-Lingual Transfer}

The results in the cross-lingual debiasing transfer are shown in Table~\ref{tbl:xling}. For brevity, we show only the results on XWEAT T8 (gender bias wrt. science vs.~art) and for a subset of evaluation measures (one for each evaluation aspect): WEAT (W), KMeans++ (KM), and SimLex (SL).\footnote{We provide the full results, with all evaluation measures, and also on the XWEAT T1 test in the supplemental material.}\textsuperscript{,}\footnote{We evaluate word similarities for \textsc{de}, \textsc{it}, \textsc{ru}, and \textsc{hr} on their respective SimLex datasets \cite{leviant2015separated,Mrksic:2017tacl}; there is no \textsc{es} and \textsc{tr} SimLex.}  

We first confirm the results from \citet{lauscher2019we}: \textsc{de}, \textsc{it}, \textsc{ru}, and \textsc{hr} fastText vectors do not exhibit significant explicit gender bias (wrt. science vs.~art), according to the WEAT test. The explicit bias is, however, significant in \textsc{es} and \textsc{tr} distributional vectors. Implicit bias is clearly present in all distributional spaces except \textsc{ru}. Debiasing models display similar properties as before: DBN reduces the explicit bias more effectively than BAM and GBDD, but it semantically distorts the vectors; and only GBDD successfully removes the implicit bias. None of the models fully removes the explicit bias for \textsc{tr} (the lowest bias effect of $0.99$ for BAM $\circ$ GBDD is still significant). We suspect that this is a result of the lower-quality cross-lingual \textsc{tr}$\rightarrow$\textsc{en} projection, which is in line with the bilingual lexicon induction results from \citet{glavas2019properly}. 

For \textsc{de} and \textsc{it}, BAM and DBN \textit{invert} the direction of the bias: negative WEAT scores mean that \textit{sciences} are more correlated with \textit{female} attributes and \textit{arts} with \textit{male} attributes. We believe that this is the result of applying a (strong) bias correction learned on a biased \textsc{en} space on the (explicitly) unbiased \textsc{de} and \textsc{it} spaces. The BAM $\circ$ GBDD composition seems most robust in the cross-lingual transfer setting -- it successfully removes both the explicit (if they exist) and implicit biases, while preserving useful semantic information (SL). These results indicate that we can attenuate or remove biases in distributional vectors of languages for which (1) we do not require the initial bias specification and (2) we do not even need similarity-specialized word embeddings used to augment the bias specifications for the target language.


\paragraph{Topology of Debiased Spaces.} 
Finally, we qualitatively analyze the debiasing effects suggested by evaluation measures. We project the input and the debiased embeddings into 2D with PCA, and show the constellation of words from the initial bias specification of WEAT T8 (Table~\ref{tbl:augmentation}) in Figure~\ref{fig:topology}.\footnote{We show only the input space and the spaces debiased with GBDD and BAM. We provide similar illustrations for other debiasing models in the supplementary material.} In the distributional space, the two target sets (\textit{science} vs \textit{art}) are clearly distinguishable from one another (implicit bias), and so are the \textit{male} and \textit{female} attributes. The \textit{science} terms are notably closer to the \textit{male} terms and \textit{art} terms to the \textit{female} terms (explicit bias). The space produced by BAM intertwines the \textit{male} and \textit{female} terms and makes the \textit{science} and \textit{art} terms roughly equidistant to the gender terms (explicit bias removed), but the \textit{science} terms are still clearly distinguishable from \textit{art} terms (implicit bias still present). In the space produced by GBDD, both biases are removed: \textit{science} and \textit{art} terms cannot be clearly separated and are roughly equidistant to gender terms.       

\section{Conclusion} 

We have introduced a general framework for debiasing distributional word vector spaces by 1) formalizing the differences between implicit and explicit biases, 2) proposing new debiasing methods that deal with the two different bias specifications, and 3) designing a comprehensive evaluation framework for testing the (often complementary) effects of debiasing. While the proposed framework offers a systematized view on human biases encoded in word embeddings, the main results indicate that our debiasing methods can effectively attenuate biases in arbitrary input distributional spaces and can also be transferred to a variety of target languages. 


\bibliography{references.bib}
\bibliographystyle{aaai}

\clearpage
\setlength{\tabcolsep}{3.7pt}
\begin{table*}[!t]
\def\arraystretch{0.85}
\centering
{
\begin{tabularx}{\linewidth}{l l ccc cc cc  ccc cc cc cc}
\toprule
& & \multicolumn{7}{c}{\textsc{DE}} & \multicolumn{7}{c}{\textsc{ES}} \\
\cmidrule(lr){3-9} \cmidrule(lr){10-16}
 & & \multicolumn{3}{c}{Explicit} & \multicolumn{2}{c}{Implicit} & \multicolumn{2}{c}{SemQ} & \multicolumn{3}{c}{Explicit} & \multicolumn{2}{c}{Implicit} & \multicolumn{2}{c}{SemQ} \\ 
\cmidrule(lr){3-5} \cmidrule(lr){6-7} \cmidrule(lr){8-9} \cmidrule(lr){10-12} \cmidrule(lr){13-14} \cmidrule(lr){15-16} 
\multicolumn{2}{c}{Model} & WEAT & ECT & BAT & KM & SVM & SL & WS & WEAT & ECT & BAT & KM & SVM & SL & WS \\ \midrule
\rowcolor{Gray}
\textbf{WEAT1} & Distributional & 1.36 & 41.7 & 59.9 & 98.9 & 75.7 & 40.7 &  68.0
& 1.47 & 61.8 & 48.1 & 100 & 57.5 & -- & --  \\ \hdashline
& GBDD & 0.42* & 77.7 & 48.2 & 90.5 & 51 & 40.7 & \textbf{68.1}
& 0.56 & \textbf{89.4} & \textbf{34.4} & 96.8 & 50.3 & -- & --  \\
\rowcolor{Gray}
& BAM & 1.39 & 50.6 & 54 & 95 & 94.3 & 39 & 64.5
& 1.12 & 62.9 & 42.2 & 97.7 & 95.3 & -- & --  \\
& DN & 0.42* & 48.1 & 48.3 & 98.9 & 53 & 39.9 & 61.9 
& 0.96 & 55.8 & 41.6 & 97.7 & 34.4 & --& -- \\ \hdashline
\rowcolor{Gray}
& GBDD $\circ$ BAM & 0.61 & 81.1 & \textbf{44.3} & 93.2 & 88.4 & 39.1 & 64.7
& 0.56 & 76.4 & 38.2 & 98.4 & 77 & -- & -- \\
& BAM $\circ$ GBDD & 0.75 & 74.3 & 52.4 & 90.8 & 50 & \textbf{40.8} & 64.9
& \textbf{0.48}* & 85.3 & 42.8 & \textbf{94.1} & 49.5 & -- & -- \\
\rowcolor{Gray}
& GBDD $\circ$ DN & \textbf{0.30*} & \textbf{82.8} & 45.7 & \textbf{86.6} & \textbf{42.9} & 39.6 & 61.9
& 0.69 & 75.1 & 38 & 96.2 & 38.3 & -- & -- \\ \midrule
\textbf{WEAT8} & Distributional & \textbf{0.05*} & 34.1 & 37.2 & 98.3 & 50 & 40.7 & 68 
& 1.16 & 67.8 & 36.4 & 99.8 & 50 & -- & -- \\ \hdashline
\rowcolor{Gray}
& GBDD & 0.15* & \textbf{85.3} & 30.5 & 55.4 & 50 & 40.7 & 67.7
& 0.41* & 70.9 & 31.1 & 60 & 50 & -- & -- \\
& BAM & -0.97  & 41.5& 33.6 & 97.4 & 100 & 40.7 & 65.8
& 0.11* & 70.9 & 34.4 & 99 & 100 & -- & --\\
\rowcolor{Gray}
& DN & -0.1* & 67.1 & 37.4 & 97.4 & 50 & 36.2 &  62 
& 0.76* & 74 & 48.1 & 100 & 50 & --& -- \\ \hdashline
& GBDD $\circ$ BAM & -0.12* & 83.2 & 35.2 & 56.3 & 50 & \textbf{40.8} & 65.6 
& \textbf{0.05*} & 83.7 & 33.1 & 58 & 50 & -- & -- \\
\rowcolor{Gray}
& BAM $\circ$ GBDD & \textbf{-0.09*} & 84.4 & \textbf{28.5} & \textbf{54.4} & 50 & 37.3 & 66.7  
& 0.11* & 85.9 & \textbf{28.1} & 56.6 & 50 & -- & -- \\
& GBDD $\circ$ DN & 0.35* & 73.4 & 35.7 & 57.6 & 50 & 35.9 & 61.1 
& 0.78* & \textbf{88.5} & 46.4 & \textbf{52.4} & 50 & -- & -- \\ 
\bottomrule
\end{tabularx}
}
\vspace{-2mm}
\caption{Complete cross-lingual debiasing transfer results for transfer to German (\textsc{de}) Spanish (\textsc{es}). Results obtained on the XWEAT \cite{lauscher2019we} T1 and T8 tests of respective languages.} 
\vspace{-1mm}
\end{table*}

\setlength{\tabcolsep}{3.7pt}
\begin{table*}[!t]
\def\arraystretch{0.85}
\centering
{
\begin{tabularx}{\linewidth}{l l ccc cc cc  ccc cc cc cc}
\toprule
& & \multicolumn{7}{c}{\textsc{IT}} & \multicolumn{7}{c}{\textsc{RU}} \\
\cmidrule(lr){3-9} \cmidrule(lr){10-16}
 & & \multicolumn{3}{c}{Explicit} & \multicolumn{2}{c}{Implicit} & \multicolumn{2}{c}{SemQ} & \multicolumn{3}{c}{Explicit} & \multicolumn{2}{c}{Implicit} & \multicolumn{2}{c}{SemQ} \\ 
\cmidrule(lr){3-5} \cmidrule(lr){6-7} \cmidrule(lr){8-9} \cmidrule(lr){10-12} \cmidrule(lr){13-14} \cmidrule(lr){15-16} 
\multicolumn{2}{c}{Model} & WEAT & ECT & BAT & KM & SVM & SL & WS & WEAT & ECT & BAT & KM & SVM & SL & WS \\ \midrule
\rowcolor{Gray}
\textbf{WEAT1} & Distributional & 1.28 & 57.7 & 57.2 & 97 & 54.8 & 29.8 & 64.2
& 1.28 & 57.6 & 43.5 & 96.7 & 54.3 & 25.6 & 59.2 \\ \hdashline
& GBDD & 0.02* & 81.8 & \textbf{44} & 77.3 & 51.1 & \textbf{29.8} & \textbf{64}
& 0.67 & 79.8 & \textbf{35.3} & 93.5 & 49.9 & \textbf{25.4} & \textbf{59}   \\
\rowcolor{Gray}
& BAM & 1.35 & 54 & 55.5 & 95.9 & 95.6 & 27.3 & 62.2
& 1.20 & 66 & 44.4 & 94.4 & 94.3 & 24.2 & 55.5 \\
& DN & 0.53 & 62.8 & 51.9 & 99.8 & 55.5 & 25.7 &  58.5
& \textbf{0.44*} & 57.7 & 42.7 & 96.5 & 56.3 & 24.3 & 52.6 \\ \hdashline
\rowcolor{Gray}
& GBDD $\circ$ BAM &  0.44* & 70.9 & 51.4 & 87.7 & 86.2 & 27.3 & 62.2
& 0.6 & \textbf{80.7} & 40.1 & 93.5 & 89 & 24.2 & 55.4 \\
& BAM $\circ$ GBDD & 0.29*  & 76.5 & 48.6 & \textbf{73.4} & 50.2 & 28.2 & 62.4 
& 0.65 & 80.2  & 37.7 & 92.8 & 49.6 & 25 & 56.3 \\
\rowcolor{Gray}
& GBDD $\circ$ DN & 0.2* & \textbf{83.5} & 48 & 88.1 & 57.6 & 25.8 & 58.3 
& \textbf{0.36*} & 75 & 40.7 & \textbf{91.1} & 52.4 & 24.1 & 52.5 \\ 
\midrule
\textbf{WEAT8} & Distributional & \textbf{0.10*} & \textbf{92.5} & 25.9 & 99.8 & 50 & 29.8 & 64.2
& 0.37* & 49.9 & 32.1 & 62 & 50 & 25.6 & 59.2 \\ \hdashline
\rowcolor{Gray}
& GBDD & -0.28*  & 86.4 & 25.9 & \textbf{56.1} & 50 & \textbf{29.8} & \textbf{63.4}
& 0.73* & 49.5 & 32 & 62.4 & 50 & \textbf{25.8} & \textbf{58.3}\\
& BAM & -0.70* & 57.4 & 23 & 99.6 & 100 & 29 & 61
& -0.41* & 44.6 & \textbf{25.9} & 74.4 & 100 & 25.1 & 56.8\\
\rowcolor{Gray}
& DN & -1.05 & 40.7 & \textbf{14.1} & 100 & 50 & 25.4 & 57.7 
& \textbf{0.31*} & 46.8 & 35.5 & 77.9 & 50 & 20.7 & 56.9 \\ \hdashline
& GBDD $\circ$ BAM & -0.62* & 67 & 23.1 & 57.9 & 50 & 29 & 60 
& 0.34* & 72.7 & 30.8 & \textbf{56.8} & 50 & 24.8 & 55.8 \\
\rowcolor{Gray}
& BAM $\circ$ GBDD & \textbf{-0.05*} & 82.3 & 28.9 & 58.9 & 50 & 27.1 &  60.2 
& 0.59* & \textbf{83.7} & 31 & 61.6 & 50 & 25.4 &  57.5\\
& GBDD $\circ$ DN & -0.64* & 51.2 & 18.7 & 60.1 & 50 & 25 & 56.7 
& 0.77*  & 69.7 & 38.3 & 61.9 & 50 & 20.7 & 55.1 \\ 
\bottomrule
\end{tabularx}
}
\vspace{-2mm}
\caption{Complete cross-lingual debiasing transfer results for transfer to Italian (\textsc{it}) and Russian (\textsc{ru}). Results obtained on the XWEAT \cite{lauscher2019we} T1 and T8 tests of respective languages.} 
\vspace{-1mm}
\end{table*}

\setlength{\tabcolsep}{5pt}
\begin{table*}[!t]
\def\arraystretch{0.85}
\centering
{
\begin{tabularx}{\linewidth}{l l ccc cc cc  ccc cc cc cc}
\toprule
& & \multicolumn{7}{c}{\textsc{HR}} & \multicolumn{7}{c}{\textsc{TR}} \\
\cmidrule(lr){3-9} \cmidrule(lr){10-16}
 & & \multicolumn{3}{c}{Explicit} & \multicolumn{2}{c}{Implicit} & \multicolumn{2}{c}{SemQ} & \multicolumn{3}{c}{Explicit} & \multicolumn{2}{c}{Implicit} & \multicolumn{2}{c}{SemQ} \\ 
\cmidrule(lr){3-5} \cmidrule(lr){6-7} \cmidrule(lr){8-9} \cmidrule(lr){10-12} \cmidrule(lr){13-14} \cmidrule(lr){15-16} 
\multicolumn{2}{c}{Model} & WEAT & ECT & BAT & KM & SVM & SL & WS & WEAT & ECT & BAT & KM & SVM & SL & WS \\ \midrule
\rowcolor{Gray}
\textbf{WEAT1} & Distributional & 1.45 & 56.3 & 63.4 & 57 & 51.7 & 32.7 & --  
& 1.21 & 69.6 & 47.9 & 86.3 & 50.6 & -- &  -- \\ \hdashline
& GBDD & 0.85 & 81.2 & 60.5 & 63.2 & 49.8 & \textbf{32.8} & --
& 0.64 & 83.9 & 40.9 & 79.7 & 51.4 & -- &  -- \\
\rowcolor{Gray}
& BAM & 1.35 & 50.8 & 63.8 & \textbf{59.5} & 90.5 & 31.2 & --
& 0.89 & 64.8 & 39.1 & 84.3 & 90.6 & -- & --  \\
& DN & 0.86 & 74.8 & 67.2 & 87.4 & 35.8 & 28.4 & -- 
& 0.78 & 73.3 & 36.9 & 88.1 & 58.3 & --& -- \\ \hdashline
\rowcolor{Gray}
& GBDD $\circ$ BAM & 0.82 & 63.6 & \textbf{57.1} & 55.1 & 77.5  & 31.3 & --
& \textbf{0.19*} & 80 & 34.5 & \textbf{72} & 73.2 & -- & -- \\
& BAM $\circ$ GBDD & 0.71 & 86.8 & 63 & 68.7 & 50 & 30.9 &--  
& 0.76 & 82.3 & 53 & 75 & 51.1  & -- & -- \\
\rowcolor{Gray}
& GBDD $\circ$ DN & 0.56* & \textbf{85.9} & 65.5 & 61.4 & \textbf{44} & 28.5 & --
& 0.63 & 81.5 & \textbf{33} & 74.7 & 54.9 & -- & -- \\ 
\midrule
\textbf{WEAT8} & Distributional & 0.13* & 53.2 & 39.4 & 98.6 & 50 & 32.7 & --
& 1.72 & 39.6 & 64.5 & 99.3 & 50 & -- & -- \\ \hdashline
\rowcolor{Gray}
& GBDD & 0.54* & 59.7 & 40.2 & \textbf{59.9} & 50 & \textbf{32.5} &-- 
& 1.41 & 71.9 & 66.5 & 64.3 & 50 & -- & -- \\
& BAM & \textbf{-0.01*} & 30.3 & 41.1 & 93.5 & 100 & 32 & --
& 1.49 & 62.1 & 59.5 & 98.8 & 100 & -- & --\\
\rowcolor{Gray}
& DN & 0.25* & 81.7 & 52.8 & 99.9 & 50 & 25.3 &--   
& 1.54 & 44.6 & 65.5 & 100 & 50 & --& -- \\ \hdashline
& GBDD $\circ$ BAM & 0.52* & 73.8 & 47 & 60.8 & 50 & 31.7 &  --
& \textbf{0.99} & 85.3 & \textbf{56} & \textbf{56.9} & 50 & -- & -- \\
\rowcolor{Gray}
& BAM $\circ$ GBDD & 0.68* & 60.9 & 44.5 & 75.4 & 50 & 29.4 & --  
& 1.27 & 59.3 & 76 & 62.4 & 50 & -- & -- \\
& GBDD $\circ$ DN & 0.67* & \textbf{88.5} & \textbf{56.6} & 67.5 & 50 & 25.1 & -- 
& 1.29 & \textbf{86.7} & 65 & 62.5 & 50 & -- & -- \\ 
\bottomrule
\end{tabularx}
}
\vspace{-2mm}
\caption{Complete cross-lingual debiasing transfer results for transfer to Croatian (\textsc{hr}) and Turkish (\textsc{tr}). Results obtained on the XWEAT \cite{lauscher2019we} T1 and T8 tests of respective languages.} 
\vspace{-1mm}
\end{table*}

\setlength{\tabcolsep}{4pt}
\begin{table*}[t!]
\centering
{\fontsize{8pt}{8pt}\selectfont
\begin{tabularx}{\linewidth}{c l lX}
\toprule
\multirow{4}{4em}{Initial} & $T_1$ & \makecell[l]{\textit{aster} \textit{clover} \textit{hyacinth} \textit{marigold} \textit{poppy} \textit{azalea} \textit{crocus} \textit{iris} \textit{orchid} \textit{rose} \textit{blue-bell} \textit{daffodil} \textit{lilac} \textit{pansy} \textit{tulip} \textit{buttercup} \textit{daisy} \textit{lily} \\ \textit{peony} \textit{violet} \textit{carnation} \textit{gladiola} \textit{magnolia} \textit{petunia} \textit{zinnia}}\\ 
& $T_2$ & \makecell[l]{\textit{ant} \textit{caterpillar} \textit{flea} \textit{locust} \textit{spider} \textit{bedbug} \textit{centipede} \textit{fly} \textit{maggot} \textit{tarantula} \textit{bee} \textit{cockroach} \textit{gnat} \textit{mosquito} \textit{termite} \textit{beetle} \textit{cricket} \\ \textit{hornet} \textit{moth} \textit{wasp} \textit{blackfly} \textit{dragonfly} \textit{horsefly} \textit{roach} \textit{weevil}}\\ 
& $A_1$ & \makecell[l]{\textit{caress} \textit{freedom} \textit{health} \textit{love} \textit{peace} \textit{cheer} \textit{friend} \textit{heaven} \textit{loyal} \textit{pleasure} \textit{diamond} \textit{gentle} \textit{honest} \textit{lucky} \textit{rainbow} \textit{diploma} \textit{gift}\\ \textit{honor} \textit{miracle} \textit{sunrise} \textit{family} \textit{happy} \textit{laughter} \textit{paradise} \textit{vacation}} \\
& $A_2$ & \makecell[l]{\textit{abuse} \textit{crash} \textit{filth} \textit{murder} \textit{sickness} \textit{accident} \textit{death} \textit{grief} \textit{poison} \textit{stink} \textit{assault} \textit{disaster} \textit{hatred} \textit{pollute} \textit{tragedy} \textit{divorce} \textit{jail} \\ \textit{poverty} \textit{ugly} \textit{cancer} \textit{kill} \textit{rotten} \textit{vomit} \textit{agony} \textit{prison}} \\ 
\midrule
\multirow{8}{4em}{k=2} & $T_1$ & \makecell[l]{\textit{glovers} \textit{gladiolus} \textit{nance} \textit{crowfoot} \textit{meadowsweet} \textit{dianthus} \textit{pinkish} \textit{dolly} \textit{poppies} \textit{cyclamen} \textit{tulips} \textit{sapphire} \textit{azaleas} \textit{wisteria} \\
\textit{camellia} \textit{asters} \textit{trefoil} \textit{sissy} \textit{olive} \textit{penstemon} \textit{candlewood} \textit{prunella} \textit{primula} \textit{mauve} \textit{opium} \textit{buddleja} \textit{taupe} \textit{magenta} \textit{veronica} \\
\textit{hyacinths} \textit{magnolias} \textit{watercress} \textit{minaj} \textit{cowslip} \textit{lilies} \textit{tulipa} \textit{orchis} \textit{daffodils} \textit{scarlet} \textit{jasmine} \textit{faggot} \textit{marigolds} \textit{orchids}}\\ 
& $T_2$ & \makecell[l]{\textit{caterpillars} \textit{gnats} \textit{termites} \textit{avenger} \textit{ants} \textit{bumblebee} \textit{arachnid} \textit{sticking} \textit{cricketing} \textit{flit} \textit{tarantulas} \textit{pyralidae} \textit{harrier} \textit{millipede} \\
\textit{centipedes} \textit{mosquitos} \textit{vermin} \textit{worm} \textit{cockroaches} \textit{locusts} \textit{wasps} \textit{insect} \textit{snook} \textit{larva} \textit{scoot} \textit{gracillariidae} \textit{weevils} \textit{grasshopper}\\
\textit{undershot} \textit{fathead} \textit{whitefly} \textit{louse} \textit{batsman} \textit{dragonflies}} \\
& $A_1$ & \makecell[l]{\textit{donation} \textit{liberty} \textit{tranquility} \textit{fortunate} \textit{mild} \textit{laugh} \textit{diamonds} \textit{holiday} \textit{truthful} \textit{endowment} \textit{untried} \textit{fitness} \textit{colleague} \\ 
\textit{credentials} \textit{lineage} \textit{gurgling} \textit{honour} \textit{faithful} \textit{cheerfulness} \textit{auspicious} \textit{affection} \textit{prism} \textit{genuine} \textit{esteem} \textit{moonlight} \textit{newfound} \\ 
\textit{vacations} \textit{gem} \textit{eden} \textit{peacefulness} \textit{gladden} \textit{wellness} \textit{partner} \textit{glad} \textit{cuddle} \textit{cherish} \textit{joy} \textit{liege} \textit{diplomas} \textit{phenomenon} \textit{fondle}\\ 
\textit{autonomy} \textit{prodigy} \textit{tickled} \textit{enjoyment} \textit{clement} \textit{utopia} \textit{tribe}}\\ 
& $A_2$ & \makecell[l]{\textit{misuse} \textit{collision} \textit{stench} \textit{destitution} \textit{demise} \textit{anguish} \textit{annihilate} \textit{estrangement} \textit{illness} \textit{incarcerate} \textit{sorrow} \textit{mistreat} \textit{infection}\\ 
\textit{destroy} \textit{separation} \textit{slaughter} \textit{antipathy} \textit{penitentiary} \textit{smash} \textit{regurgitate} \textit{malady} \textit{misery} \textit{decease} \textit{dirt} \textit{calamity} \\
\textit{impoverishment} \textit{spew} \textit{stinking} \textit{toxin} \textit{enmity} \textit{imprison} \textit{tainted} \textit{massacre} \textit{gaol} \textit{sinister} \textit{horrible} \textit{defile} \textit{contaminate} \textit{reek} \\
\textit{prostate} \textit{catastrophe} \textit{crud} \textit{casualty} \textit{mishap} \textit{leukemia} \textit{invasion} \textit{misadventure} \textit{onslaught}}\\
\midrule
\multirow{8}{4em}{k=3} & $T_1$ & \makecell[l]{\textit{faggot} \textit{cornflower} \textit{meadowsweet} \textit{cowslip} \textit{camellia} \textit{cress} \textit{weeknd} \textit{orchidaceae} \textit{watercress} \textit{trefoil} \textit{pinkish} \textit{magnoliaceae} \\
\textit{orchids} \textit{lilies} \textit{dianthus} \textit{hyacinths} \textit{primula} \textit{willowherb} \textit{daffodils} \textit{mauve} \textit{penstemon} \textit{azaleas} \textit{fleabane} \textit{magenta} \textit{wisteria} \textit{jessie}\\
\textit{licorice} \textit{lilacs} \textit{polly} \textit{peonies} \textit{magnolias} \textit{candlewood} \textit{amaranthus} \textit{jasmine} \textit{opium} \textit{bluish} \textit{poppies} \textit{sapphire} \textit{orchis} \textit{sissy}\\
 \textit{buddleja} \textit{tangerine} \textit{olive} \textit{clovers} \textit{marigolds} \textit{lavender} \textit{dandelions} \textit{tulipa} \textit{taupe} \textit{tulips} \textit{poof} \textit{crowfoot} \textit{gladiolus} \textit{prunella} \\
\textit{dandelion} \textit{veronica} \textit{dolly} \textit{asters} \textit{cyclamen} \textit{scarlet} \textit{minaj} \textit{nance}} \\
& $T_2$ & \makecell[l]{\textit{projected} \textit{avenger} \textit{grasshopper} \textit{vermin} \textit{scamper} \textit{worm} \textit{cockroaches} \textit{fathead} \textit{harrier} \textit{batsman} \textit{weevils} \textit{snook} \textit{whitefly} \textit{bug} \\
 \textit{noctuidae} \textit{scorpion} \textit{mayfly} \textit{tarantulas} \textit{louse} \textit{roaches} \textit{cricketing} \textit{bumblebee} \textit{gnats} \textit{curculionidae} \textit{arachnid} \textit{mosquitoes} \textit{wasps}\\
\textit{dragonflies} \textit{scoot} \textit{termites} \textit{larva} \textit{millipede} \textit{corsair} \textit{flit} \textit{gracillariidae} \textit{locusts} \textit{wicket} \textit{hive} \textit{insect} \textit{caterpillars} \textit{mosquitos}\\
\textit{parasitoid} \textit{undershot} \textit{sticking} \textit{centipedes} \textit{ants} \textit{pyralidae} \textit{fleas}}\\
& $A_1$ & \makecell[l]{\textit{fortunate} \textit{colleague} \textit{auspicious} \textit{peacefulness} \textit{untried} \textit{jewel} \textit{propitious} \textit{cherish} \textit{joy} \textit{truthful} \textit{stunner} \textit{hug} \textit{dearest} \textit{partner}\\
\textit{comrade} \textit{honour} \textit{gladden} \textit{glad} \textit{bliss} \textit{delight} \textit{encourage} \textit{mild} \textit{eden} \textit{laugh} \textit{moonlight} \textit{genuine} \textit{tickled} \textit{joyful} \textit{diamonds} \textit{gem}\\ 
\textit{gratuity} \textit{sabbatical} \textit{enjoyment} \textit{lineage} \textit{endowment} \textit{liberty} \textit{certificate} \textit{newfound} \textit{liege} \textit{wellness} \textit{gurgling} \textit{credentials} \textit{clement}\\ 
\textit{utopia} \textit{autonomy} \textit{faithful} \textit{tribe} \textit{chuckle} \textit{vacations} \textit{prism} \textit{holiday} \textit{serenity} \textit{sincere} \textit{phenomenon} \textit{diplomas} \textit{homage} \textit{rainbows}\\
\textit{donation} \textit{cuddle} \textit{welfare} \textit{tranquility} \textit{affection} \textit{allegiant} \textit{independency} \textit{tranquil} \textit{prodigy} \textit{esteem} \textit{fondle} \textit{cheerfulness} \textit{ancestry}\\
\textit{fitness} \textit{untested}}\\
& $A_2$ & \makecell[l]{\textit{severance} \textit{reek} \textit{imprison} \textit{onslaught} \textit{surly} \textit{destroy} \textit{massacre} \textit{invasion} \textit{complaint} \textit{spew} \textit{dirt} \textit{casualty} \textit{heartbreak} \textit{slaying} \\
\textit{stinking} \textit{catastrophe} \textit{penitentiary} \textit{demise} \textit{slaughter} \textit{privation} \textit{toxin} \textit{illness} \textit{impoverishment} \textit{annihilate} \textit{calamity} \\
\textit{contaminate} \textit{separation} \textit{collision} \textit{outrage} \textit{grime} \textit{stench} \textit{disgorge} \textit{mishap} \textit{collide} \textit{hate} \textit{regurgitate} \textit{crud} \textit{misuse} \textit{malady}\\
\textit{contagion} \textit{sinister} \textit{infection} \textit{smash} \textit{attack} \textit{leukemia} \textit{tumour} \textit{tainted} \textit{anguish} \textit{defile} \textit{stinky} \textit{ailment} \textit{gaol} \textit{decease} \textit{extinguish}\\
\textit{enmity} \textit{sorrow} \textit{misadventure} \textit{expiration} \textit{pollutes} \textit{antipathy} \textit{estrangement} \textit{misery} \textit{incarcerate} \textit{horrible} \textit{prostate} \textit{destitution}\\
\textit{mistreat}}\\
\midrule
\multirow{8}{4em}{k=4} & $T_1$ & \makecell[l]{\textit{scarlet} \textit{bluebell} \textit{cornflower} \textit{delphinium} \textit{fleabane} \textit{amaranthus} \textit{dianthus} \textit{chromatic} \textit{poof} \textit{peonies} \textit{orchidaceae} \textit{orchis} \textit{azaleas}\\
\textit{mauve} \textit{tangerine} \textit{nance} \textit{tulipa} \textit{camellia} \textit{taupe} \textit{willowherb} \textit{hyacinths} \textit{minaj} \textit{periwinkle} \textit{helianthemum} \textit{poppies} \textit{lilies} \textit{cress}\\
\textit{magnolias} \textit{macklemore} \textit{dolly} \textit{sissy} \textit{sapphire} \textit{orchids} \textit{buddleja} \textit{licorice} \textit{jasmine} \textit{faggot} \textit{tulips} \textit{lavender} \textit{opium} \textit{dandelion}\\ 
\textit{weeknd} \textit{wisteria} \textit{cowslip} \textit{prunella} \textit{thyme} \textit{alfalfa} \textit{lilacs} \textit{daffodils} \textit{magnoliaceae} \textit{pinkish} \textit{watercress} \textit{crowfoot} \textit{veronica} \\
\textit{primula} \textit{carrie} \textit{bluish} \textit{cryptanthus} \textit{trefoil} \textit{asters} \textit{jessie} \textit{polly} \textit{olive} \textit{clovers} \textit{meadowsweet} \textit{fuchsia} \textit{penstemon} \textit{candlewood} \\
\textit{marigolds} \textit{dandelions} \textit{cyclamen} \textit{snowberry} \textit{purplish} \textit{sassafras} \textit{gladiolus} \textit{epiphyte} \textit{magenta}} \\
& $T_2$ & \makecell[l]{\textit{caterpillars} \textit{wasps} \textit{corsair} \textit{whitefly} \textit{insect} \textit{bumblebee} \textit{bowler} \textit{noctuidae} \textit{yellowjacket} \textit{mayfly} \textit{curculionidae} \textit{cockroaches} \\
 \textit{dragonflies} \textit{avenger} \textit{mulligan} \textit{pilotless} \textit{roundworm} \textit{undershot} \textit{protruding} \textit{grasshopper} \textit{crambidae} \textit{damselfly} \textit{louse} \\
\textit{projected} \textit{cricketing} \textit{vermin} \textit{parasitoid} \textit{tarantulas} \textit{wicket} \textit{sticking} \textit{scorpion} \textit{gnats} \textit{hellcat} \textit{mosquitoes} \textit{sawfly} \textit{hive} \textit{arachnid} \\
\textit{larva} \textit{locusts} \textit{centipedes} \textit{snook} \textit{batsman} \textit{weevils} \textit{dart} \textit{flit} \textit{bug} \textit{fleas} \textit{gracillariidae} \textit{harrier} \textit{burrowing} \textit{scamper} \textit{roaches} \\
\textit{hickory} \textit{mosquitos} \textit{scoot} \textit{tractor} \textit{fathead} \textit{worm} \textit{bumblebees} \textit{millipede} \textit{pyralidae} \textit{termites} \textit{leafhopper} \textit{ants}}\\
& $A_1$ & \makecell[l]{\textit{independency} \textit{rhombus} \textit{daybreak} \textit{endowment} \textit{enliven} \textit{vacationing} \textit{cheerful} \textit{tribe} \textit{partner} \textit{privilege} \textit{truthful} \textit{rainbows} \textit{gem} \\ \textit{gratification} \textit{gratuity} \textit{affection} \textit{phenomenon} \textit{delight} \textit{untried} \textit{daydream} \textit{mirth} \textit{fondle} \textit{tranquility} \textit{prism} \textit{gladden} \textit{enjoyment} \\ \textit{esteem} \textit{stunner} \textit{certificate} \textit{genuine} \textit{holiday} \textit{glad} \textit{sabbatical} \textit{encourage} \textit{autonomy} \textit{cherish} \textit{baccalaureate} \textit{favorable} \\ \textit{credentials} \textit{donation} \textit{tranquil} \textit{fitness} \textit{wellness} \textit{mild} \textit{reverence} \textit{hug} \textit{benefaction} \textit{gracious} \textit{diplomas} \textit{ancestry} \textit{nirvana} \textit{staunch} \\ \textit{chuckle} \textit{vacations} \textit{cuddle} \textit{marvel} \textit{propitious} \textit{liege} \textit{gurgling} \textit{serenity} \textit{peacefulness} \textit{honour} \textit{kiss} \textit{allegiant} \textit{utopia} \textit{welfare} \\ \textit{sincere} \textit{clement} \textit{jewel} \textit{eden} \textit{fortunate} \textit{faithful} \textit{joyful} \textit{prodigy} \textit{moonlight} \textit{homage} \textit{diamonds} \textit{tickled} \textit{laugh} \textit{dearest} \textit{sidekick} \\ \textit{colleague} \textit{untested} \textit{bliss} \textit{cheerfulness} \textit{lineage} \textit{liberty} \textit{parentage} \textit{idolize} \textit{calmness} \textit{authentic} \textit{comrade} \textit{joy} \textit{auspicious} \\
 \textit{newfound} \textit{wellbeing}} \\
& $A_2$ & \makecell[l]{\textit{stinky} \textit{protest} \textit{mistreat} \textit{sorrow} \textit{disease} \textit{maltreatment} \textit{taint} \textit{remand} \textit{horrible} \textit{casualty} \textit{contaminate} \textit{smash} \textit{misery} \textit{misuse} \\
\textit{annihilate} \textit{imprison} \textit{crud} \textit{raid} \textit{grime} \textit{pollutes} \textit{contagion} \textit{barf} \textit{infection} \textit{hate} \textit{decease} \textit{slaughter} \textit{destroy} \textit{calamity} \textit{sinister} \\
 \textit{breakup} \textit{expiration} \textit{enmity} \textit{carnage} \textit{hideous} \textit{demise} \textit{regurgitate} \textit{stench} \textit{tainted} \textit{outrage} \textit{stockade} \textit{dying} \textit{separation} \textit{invasion} \\
\textit{shatter} \textit{antipathy} \textit{happening} \textit{extinguish} \textit{privation} \textit{spew} \textit{tumour} \textit{ailment} \textit{complaint} \textit{attack} \textit{destitution} \textit{exterminate} \textit{rancid} \\
\textit{massacre} \textit{impoverishment} \textit{slaying} \textit{heartache} \textit{misfortune} \textit{incarcerate} \textit{disgorge} \textit{surly} \textit{malady} \textit{catastrophe} \textit{onslaught} \textit{collide} \\
\textit{misadventure} \textit{defile} \textit{gaol} \textit{prostate} \textit{dirt} \textit{penitentiary} \textit{anguish} \textit{dearth} \textit{animosity} \textit{muck} \textit{heartbreak} \textit{reek} \textit{severance} \\ 
\textit{contamination} \textit{collision} \textit{estrangement} \textit{illness} \textit{leukemia} \textit{tumor} \textit{mishap} \textit{toxin} \textit{stinking}} \\
\bottomrule
\end{tabularx}
}
\vspace{-2mm}
\caption{Bias specification of WEAT T1: sentiment attached to flowers ($T_1$) vs. insects ($T_2$). Original terms from \citet{Caliskan183} and augmented list for different $k$.}
\label{tbl:weat_1}
\vspace{-2mm}
\end{table*}

\setlength{\tabcolsep}{4pt}
\begin{table*}[t!]
\centering
{\fontsize{8pt}{8pt}\selectfont
\begin{tabularx}{\linewidth}{c l lX}
\toprule
\multirow{4}{4em}{Initial} & $T_1$ & \textit{science} \textit{technology} \textit{physics} \textit{chemistry} \textit{Einstein} \textit{NASA} \textit{experiment} \textit{astronomy}\\ 
& $T_2$ & \makecell[l]{\textit{poetry} \textit{art} \textit{Shakespeare} \textit{dance} \textit{literature} \textit{novel} \textit{symphony} \textit{drama}}\\
& $A_1$ & \makecell[l]{\textit{brother} \textit{father} \textit{uncle} \textit{grandfather} \textit{son} \textit{he} \textit{his} \textit{him}} \\
& $A_2$ & \makecell[l]{\textit{sister} \textit{mother} \textit{aunt} \textit{grandmother} \textit{daughter} \textit{she} \textit{hers} \textit{her}} \\
\midrule
\multirow{5}{4em}{k=2} & $T_1$ & \makecell[l]{\textit{automation} \textit{radiochemistry} \textit{test} \textit{biophysics} \textit{learning} \textit{electrodynamics} \textit{biochemistry} \textit{astrophysics} \textit{erudition} \textit{astrometry} \\ \textit{technologies} \textit{experimentation}} \\ 
& $T_2$ &  \makecell[l]{\textit{orchestra} \textit{artistry} \textit{dramaturgy} \textit{poesy} \textit{philharmonic} \textit{craft} \textit{untried} \textit{hop} \textit{poem} \textit{dancing} \textit{dissertation} \textit{treatise} \textit{new} \textit{dramatics}}\\ 
& $A_1$ & \makecell[l]{\textit{beget} \textit{buddy} \textit{forefather} \textit{man} \textit{nephew} \textit{own} \textit{himself} \textit{theirs} \textit{boy} \textit{helium} \textit{crony} \textit{cousin} \textit{grandpa} \textit{granddad} \textit{herself}} \\
& $A_2$ & \makecell[l]{\textit{niece} \textit{girl} \textit{parent} \textit{grandma} \textit{granny} \textit{woman} \textit{theirs} \textit{sire} \textit{auntie} \textit{sibling} \textit{herself} \textit{jealously} \textit{stepmother} \textit{wife}} \\
\midrule
\multirow{5}{4em}{k=3} & $T_1$ & \makecell[l]{\textit{technologies} \textit{biochemistry} \textit{astrophysics} \textit{engineering} \textit{electrodynamics} \textit{radiochemistry} \textit{astronomer} \textit{erudition} \textit{education}\\ \textit{automation} \textit{biophysics} \textit{chromodynamics} \textit{research} \textit{learning} \textit{experimentation} \textit{test} \textit{astrometry} \textit{biology}} \\
& $T_2$ & \makecell[l]{\textit{groundbreaking} \textit{craftsmanship} \textit{dissertation} \textit{new} \textit{literatures} \textit{dramatization} \textit{philharmonic} \textit{sinfonietta} \textit{artistry} \textit{untried} \\ \textit{poems} \textit{dramaturgy} \textit{dancing} \textit{dramatics} \textit{poem} \textit{poesy} \textit{craft} \textit{hop} \textit{treatise} \textit{orchestra} \textit{waltz}}\\
& $A_1$ & \makecell[l]{\textit{granddad} \textit{granddaddy} \textit{man} \textit{helium} \textit{grandpa} \textit{own} \textit{himself} \textit{forefather} \textit{themself} \textit{kinsman} \textit{theirs} \textit{sire} \textit{beget} \textit{boy} \textit{buddy} \textit{herself} \\ \textit{comrade} \textit{who} \textit{crony} \textit{nephew} \textit{grandson} \textit{cousin}} \\
& $A_2$ & \makecell[l]{\textit{sire} \textit{beget} \textit{stepmother} \textit{aunty} \textit{parent} \textit{woman} \textit{grandma} \textit{herself} \textit{own} \textit{stepsister} \textit{female} \textit{girl} \textit{jealously} \textit{sibling} \textit{auntie} \textit{theirs} \\ \textit{granny} \textit{niece} \textit{wife}} \\
\midrule
\multirow{5}{4em}{k=4} & $T_1$ & \makecell[l]{\textit{physicists} \textit{test} \textit{electrochemistry} \textit{automation} \textit{engineering} \textit{biophysics} \textit{education} \textit{learning} \textit{chromodynamics} \textit{technologies} \\ \textit{radiochemistry} \textit{examination} \textit{biology} \textit{technological} \textit{astronomer} \textit{astrophysics} \textit{experimentation} \textit{biochemistry} \textit{research} \textit{lore} \\ \textit{electrodynamics} \textit{astrobiology} \textit{astrometry} \textit{erudition}}\\
& $T_2$ & \makecell[l]{\textit{dramaturgy} \textit{monograph} \textit{untried} \textit{dances} \textit{poesy} \textit{dissertation} \textit{craftsmanship} \textit{orchestra} \textit{treatise} \textit{skill} \textit{waltz} \textit{poem} \textit{literatures}\\ \textit{dramatization} \textit{poems} \textit{theatre} \textit{dancing} \textit{newfound} \textit{hop} \textit{artistry} \textit{new} \textit{verse} \textit{craft} \textit{philharmonic} \textit{concerto} \textit{groundbreaking}\\ \textit{dramatics} \textit{sinfonietta}} \\
& $A_1$ & \makecell[l]{\textit{grandad} \textit{theirs} \textit{grandson} \textit{buddy} \textit{themself} \textit{stepbrother} \textit{forefather} \textit{ironically} \textit{crony} \textit{granddaddy} \textit{grandpa} \textit{sidekick} \textit{boy} \textit{heir}\\ \textit{granddad} \textit{cousin} \textit{who} \textit{male} \textit{man} \textit{sire} \textit{parent} \textit{beget} \textit{kinsman} \textit{nephew} \textit{herself} \textit{own} \textit{comrade} \textit{himself} \textit{helium}}\\
& $A_2$ & \makecell[l]{\textit{auntie} \textit{fiance} \textit{theirs} \textit{female} \textit{stepmother} \textit{grandma} \textit{woman} \textit{procreate} \textit{stepsister} \textit{widow} \textit{aunty} \textit{grandmothers} \textit{mimi} \textit{granny} \\ \textit{sibling} \textit{wife} \textit{sire} \textit{parent} \textit{beget} \textit{niece} \textit{herself} \textit{own} \textit{girl} \textit{jealously} \textit{siblings}} \\
\midrule
\multirow{5}{4em}{k=5} & $T_1$ & \makecell[l]{\textit{experimentation} \textit{lore} \textit{research} \textit{chromodynamics} \textit{astrobiology} \textit{technological} \textit{technologies} \textit{physicists} \textit{education} \textit{investigation}\\ \textit{engineering} \textit{examination} \textit{radiochemistry} \textit{biology} \textit{astrophysics} \textit{astrology} \textit{chemistries} \textit{learning} \textit{biochemistry}\\ \textit{electrochemistry} \textit{biophysics} \textit{astronomer} \textit{test} \textit{scholarship} \textit{electrodynamics} \textit{biotechnology} \textit{erudition} \textit{automation} \textit{astrometry} } \\
& $T_2$ & \makecell[l]{\textit{new} \textit{untried} \textit{literatures} \textit{rhyme} \textit{sinfonietta} \textit{monograph} \textit{philharmonic} \textit{hop} \textit{expertise} \textit{craft} \textit{dancing} \textit{theater} \textit{dances} \textit{newfound}\\ \textit{artistry} \textit{dramatics} \textit{untested} \textit{writing} \textit{orchestra} \textit{dramatization} \textit{poesy} \textit{craftsmanship} \textit{dramaturgy} \textit{jitterbug} \textit{theatre} \textit{treatise}\\ \textit{concerto} \textit{poem} \textit{orchestral} \textit{verse} \textit{poems} \textit{waltz} \textit{dissertation} \textit{groundbreaking} \textit{skill}} \\
& $A_1$ & \makecell[l]{\textit{granddad} \textit{crony} \textit{its} \textit{granddaddy} \textit{male} \textit{helium} \textit{herself} \textit{forefather} \textit{heir} \textit{granduncle} \textit{own} \textit{sidekick} \textit{grandson} \textit{comrade}\\ \textit{grandfathers} \textit{sire} \textit{nephew} \textit{man} \textit{stepbrother} \textit{grandad} \textit{theirs} \textit{cousin} \textit{who} \textit{hesitates} \textit{themself} \textit{parent} \textit{grandpa} \textit{kinsman}\\ \textit{ironically} \textit{himself} \textit{boy} \textit{buddy} \textit{spawn} \textit{beget} } \\
& $A_2$ & \makecell[l]{\textit{female} \textit{wife} \textit{kinswoman} \textit{girl} \textit{herself} \textit{stepsisters} \textit{stepsister} \textit{grandmothers} \textit{own} \textit{granny} \textit{stepmother} \textit{affections} \textit{woman} \textit{sire} \\ \textit{spouse} \textit{lady} \textit{theirs} \textit{fiance} \textit{aunty} \textit{procreate} \textit{progenitor} \textit{parent} \textit{jealously} \textit{sisters} \textit{siblings} \textit{niece} \textit{widow} \textit{mimi} \textit{auntie} \textit{matriarch}\\ \textit{sibling} \textit{grandma} \textit{beget}}\\
\bottomrule
\end{tabularx}
}
\vspace{-2mm}
\caption{Bias specification of WEAT T8: female vs. male attributes attached to science ($T_1$) vs. art ($T_2$). Original terms from \citet{Caliskan183} and augmented list for different $k$.}
\label{tbl:weat_8_full}
\vspace{-2mm}
\end{table*}

\clearpage
\begin{figure}[t]
    \centering
    \includegraphics[width=1.0\linewidth]{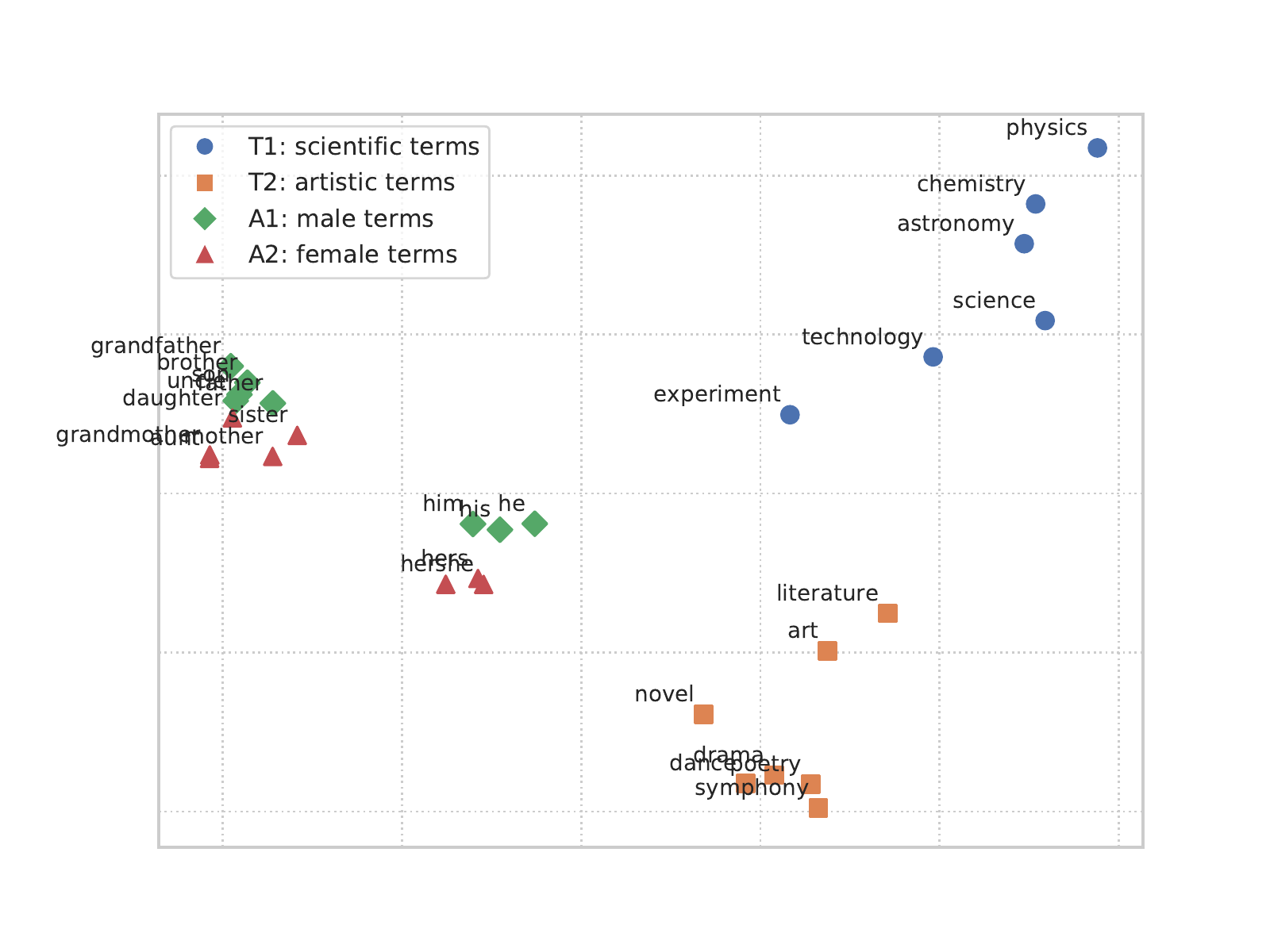}
    \vspace{-1.5mm}
    \caption{The topology of a vector space before debiasing.  Terms from WEAT T8 test: $T_1$ -- \textit{science} terms (blue), $T_2$ -- \textit{art} terms (orange), $A_1$ -- \textit{male} terms (green), and $A_2$ -- \textit{femal}e terms (red). 2D projection with PCA.}
    \label{fig:topology-distr}
    \vspace{-2.5mm}
\end{figure}

\begin{figure}[t]
    \centering
    \includegraphics[width=1.0\linewidth]{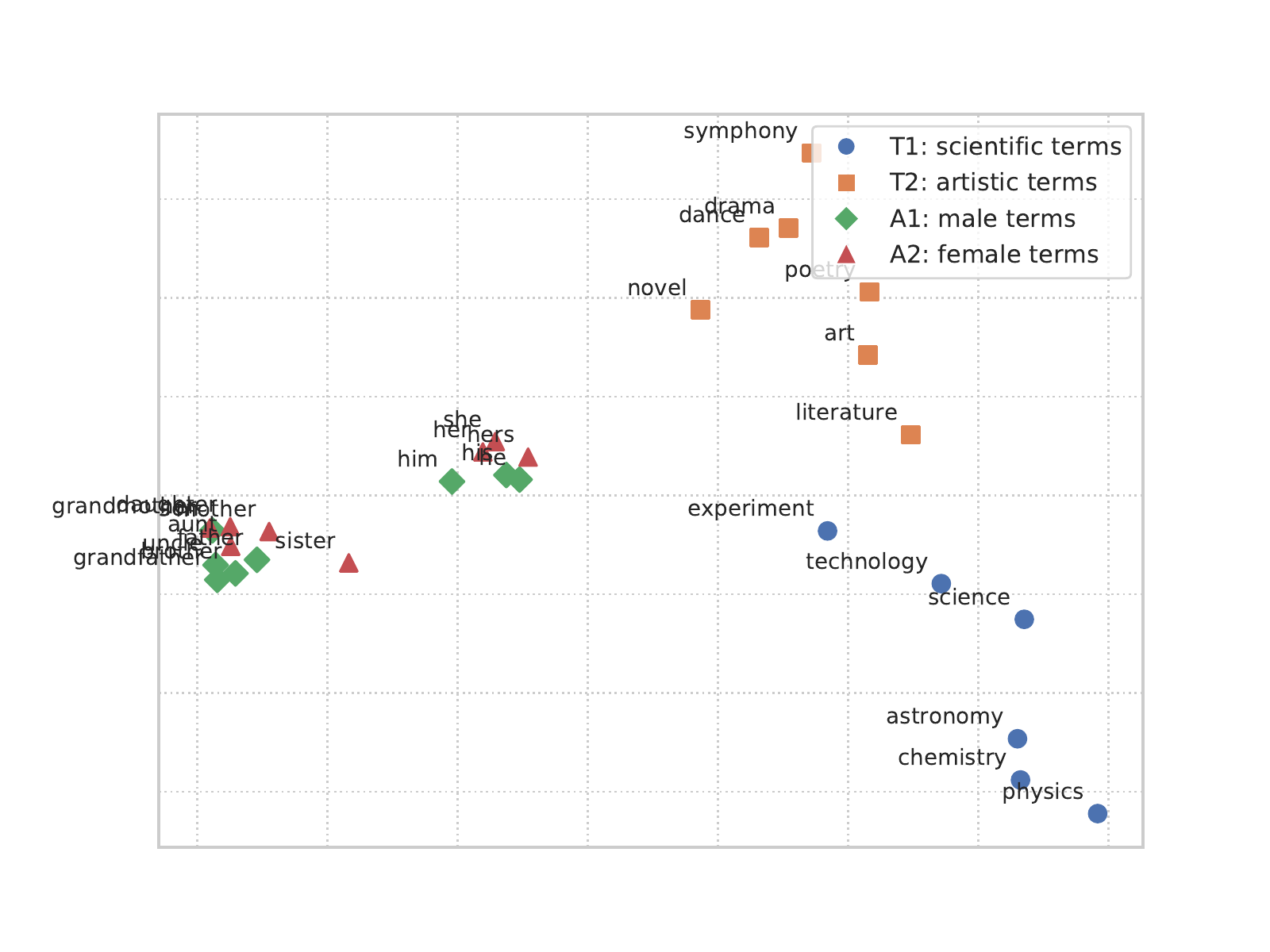}
    \vspace{-1.5mm}
    \caption{The topology of a vector space after debiasing.  Terms from WEAT T8 test: $T_1$ -- \textit{science} terms (blue), $T_2$ -- \textit{art} terms (orange), $A_1$ -- \textit{male} terms (green), and $A_2$ -- \textit{femal}e terms (red). 2D projection with PCA. Debiased using BAM.}
    \label{fig:topology-procrustes}
    \vspace{-2.5mm}
\end{figure}

\begin{figure}[t]
    \centering
    \includegraphics[width=1.0\linewidth]{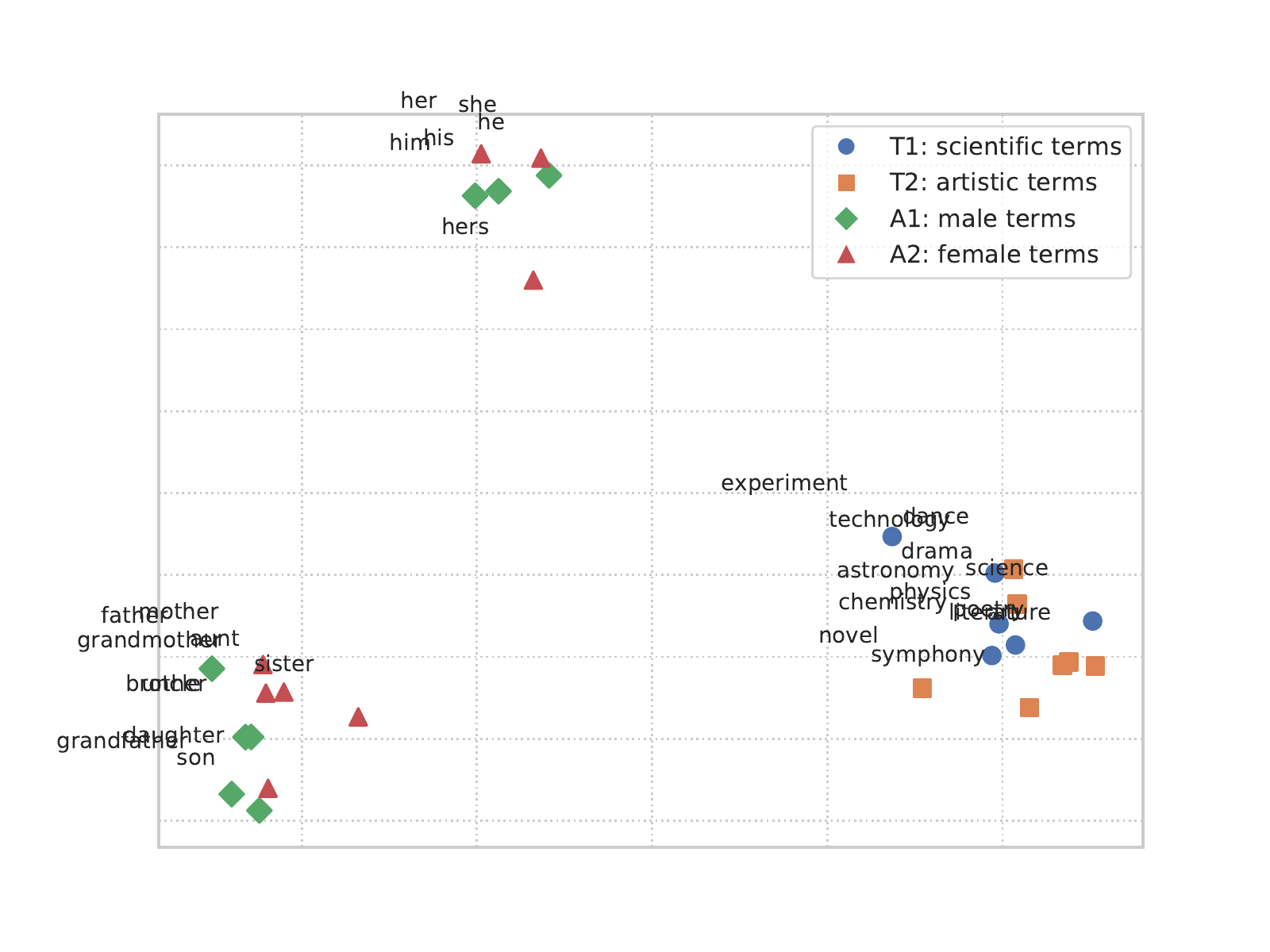}
    \vspace{-1.5mm}
    \caption{The topology of a vector space after debiasing.  Terms from WEAT T8 test: $T_1$ -- \textit{science} terms (blue), $T_2$ -- \textit{art} terms (orange), $A_1$ -- \textit{male} terms (green), and $A_2$ -- \textit{femal}e terms (red). 2D projection with PCA. Debiased using GBDD.}
    \label{fig:topology-utah}
    \vspace{-2.5mm}
\end{figure}

\begin{figure}[t]
    \centering
    \includegraphics[width=1.0\linewidth]{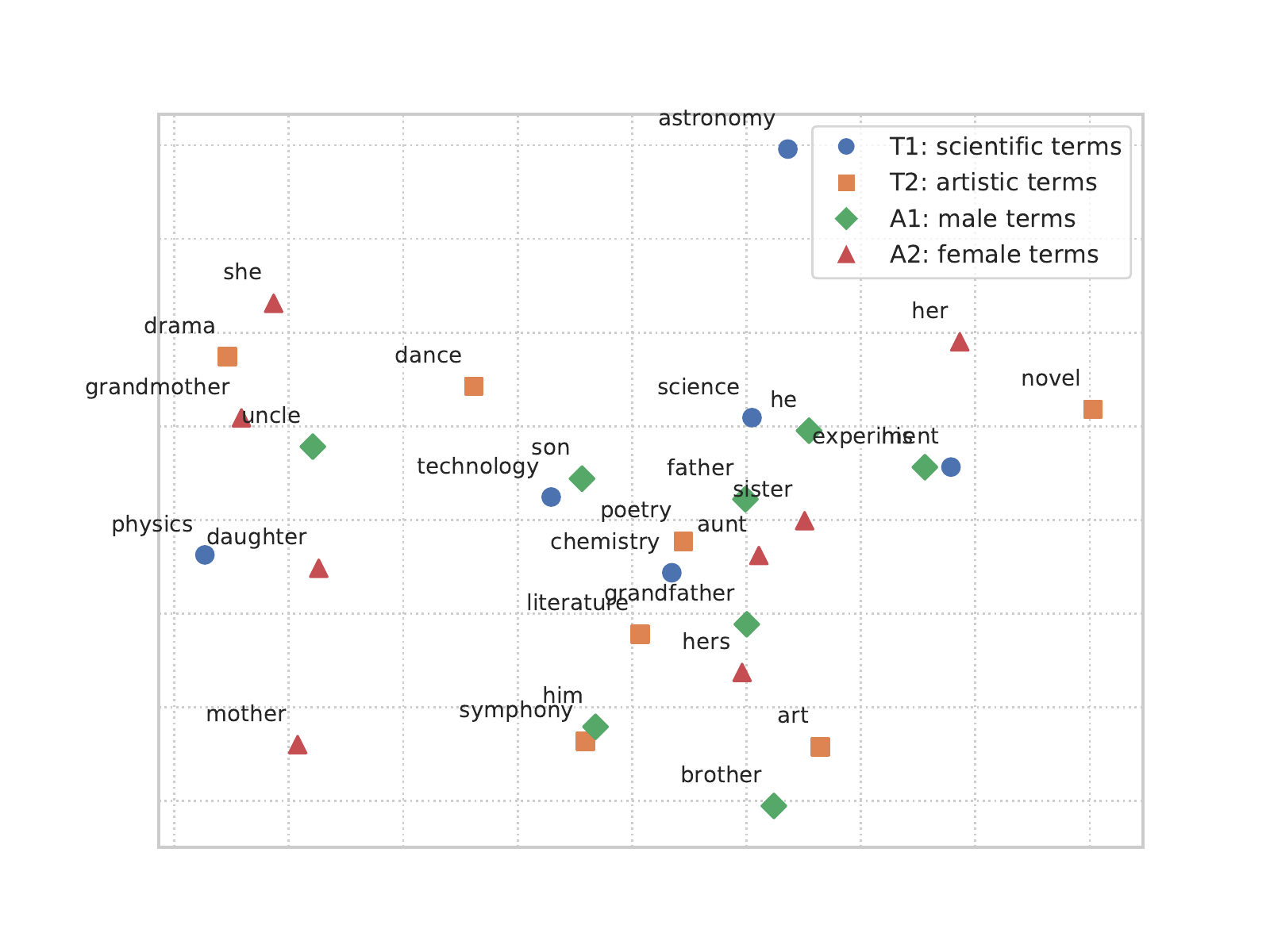}
    \vspace{-1.5mm}
    \caption{The topology of a vector space after debiasing.  Terms from WEAT T8 test: $T_1$ -- \textit{science} terms (blue), $T_2$ -- \textit{art} terms (orange), $A_1$ -- \textit{male} terms (green), and $A_2$ -- \textit{femal}e terms (red). 2D projection with PCA. Debiased using DN.}
    \label{fig:topology-anne}
    \vspace{-2.5mm}
\end{figure}

\begin{figure}[t]
    \centering
    \includegraphics[width=1.0\linewidth]{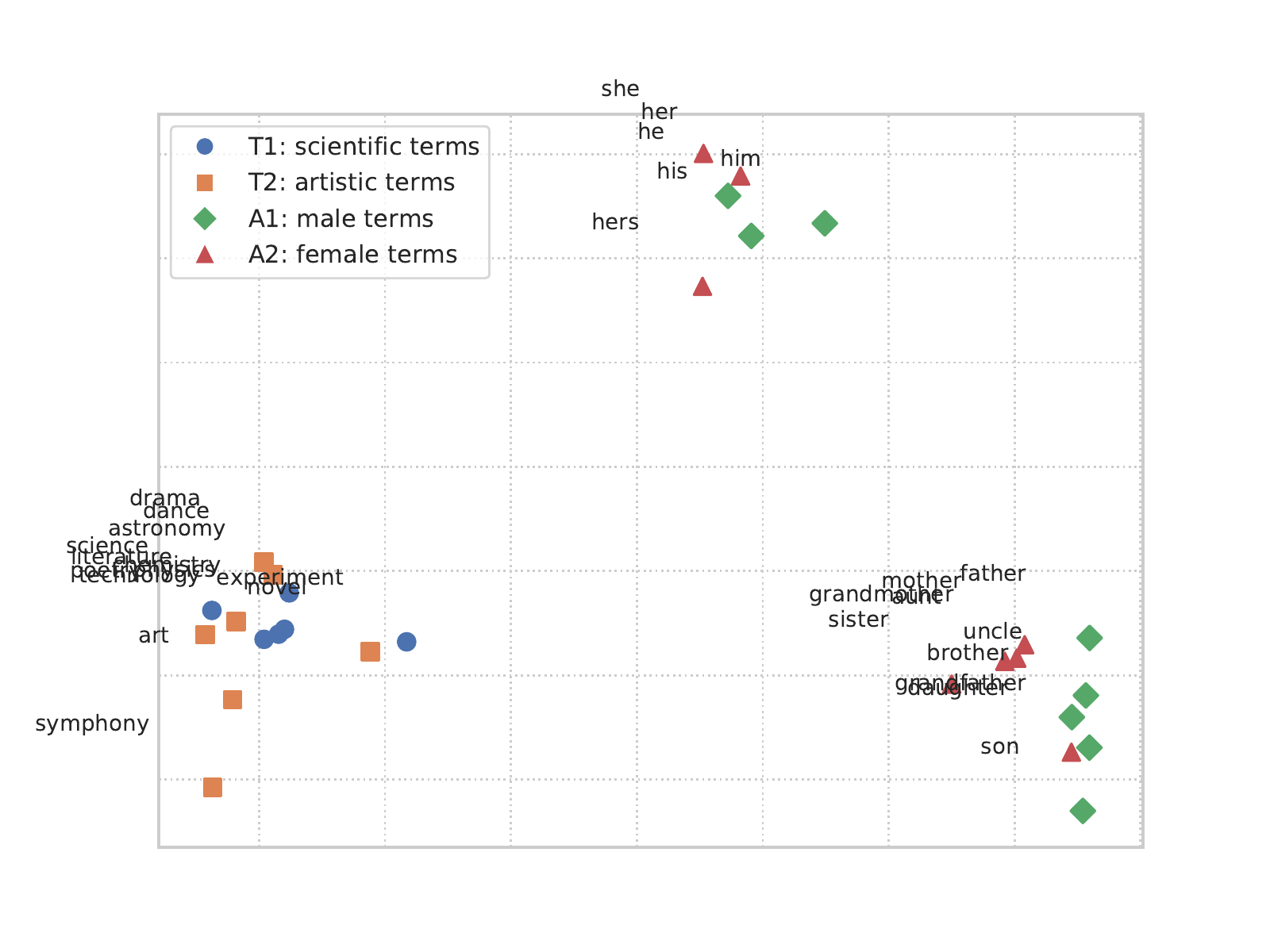}
    \vspace{-1.5mm}
    \caption{The topology of a vector space  after debiasing.  Terms from WEAT T8 test: $T_1$ -- \textit{science} terms (blue), $T_2$ -- \textit{art} terms (orange), $A_1$ -- \textit{male} terms (green), and $A_2$ -- \textit{femal}e terms (red). 2D projection with PCA. Debiased using GBDD $\circ$ BAM .}
    \label{fig:topology-utah-procrustes}
    \vspace{-2.5mm}
\end{figure}

\begin{figure}[t]
    \centering
    \includegraphics[width=1.0\linewidth]{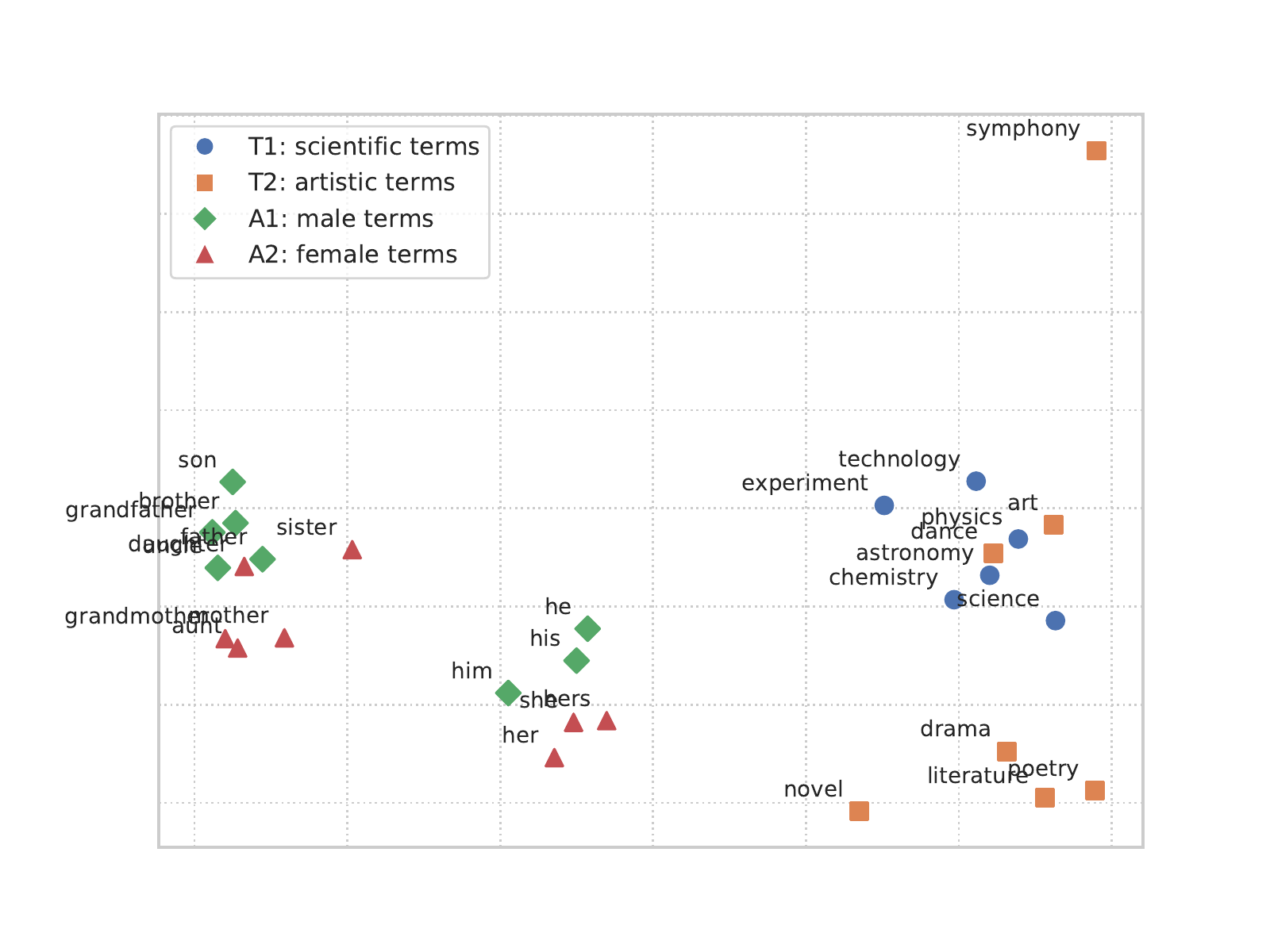}
    \vspace{-1.5mm}
    \caption{The topology of a vector space after debiasing.  Terms from WEAT T8 test: $T_1$ -- \textit{science} terms (blue), $T_2$ -- \textit{art} terms (orange), $A_1$ -- \textit{male} terms (green), and $A_2$ -- \textit{femal}e terms (red). 2D projection with PCA. Debiased using BAM $\circ$ GBDD .}
    \label{fig:topology-procrustes-utah}
    \vspace{-2.5mm}
\end{figure}

\begin{figure}[t]
    \centering
    \includegraphics[width=1.0\linewidth]{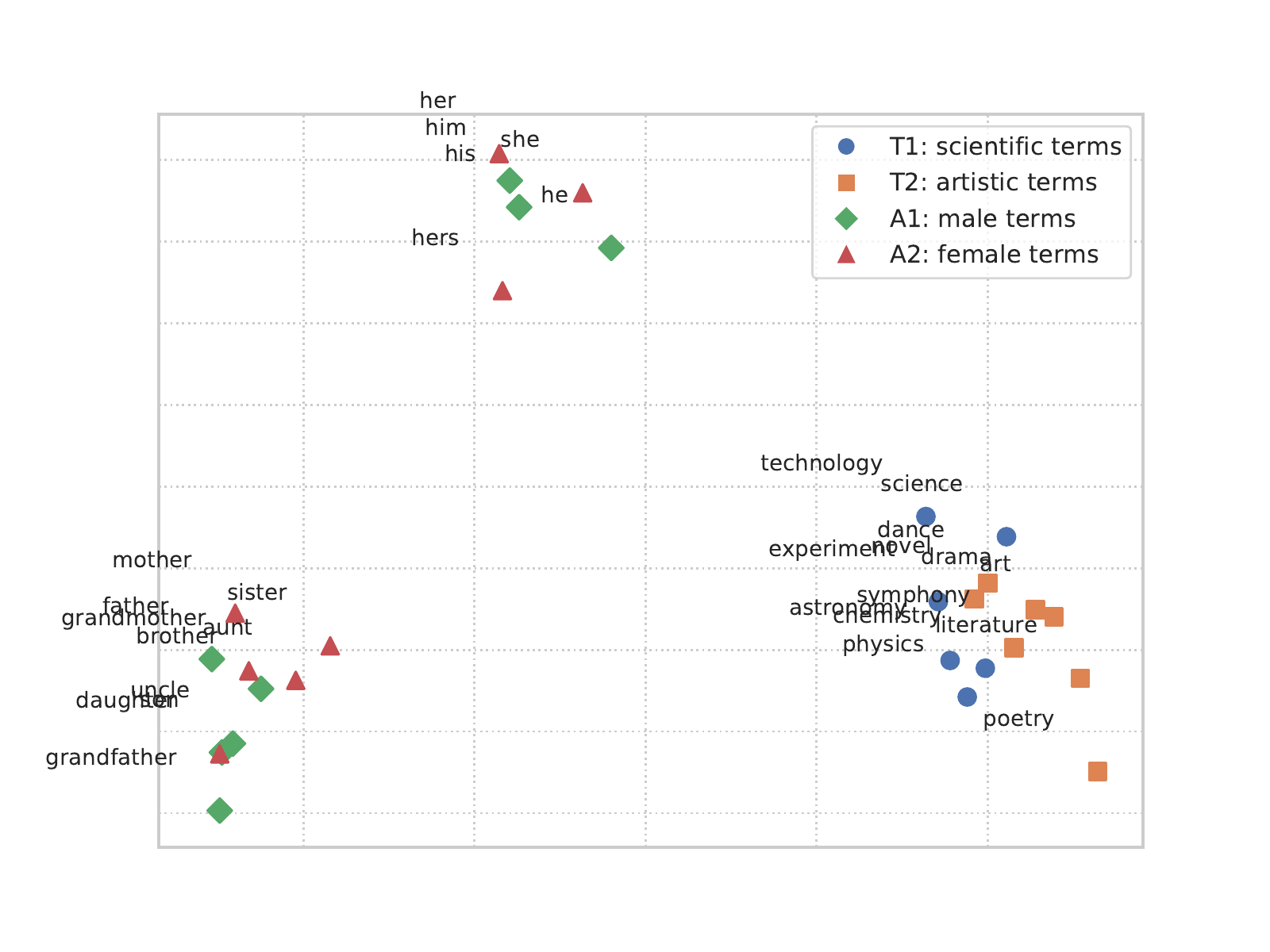}
    \vspace{-1.5mm}
    \caption{The topology of a vector space after debiasing.  Terms from WEAT T8 test: $T_1$ -- \textit{science} terms (blue), $T_2$ -- \textit{art} terms (orange), $A_1$ -- \textit{male} terms (green), and $A_2$ -- \textit{femal}e terms (red). 2D projection with PCA. Debiased using GBDD $\circ$ DN.}
    \label{fig:topology-utah-anne}
    \vspace{-2.5mm}
\end{figure}

\end{document}